\pdfoutput=1

\documentclass[11pt]{article}

\usepackage[]{acl}

\usepackage{times}
\usepackage{latexsym}

\usepackage[T1]{fontenc}

\usepackage[utf8]{inputenc}

\usepackage{microtype}

\usepackage{inconsolata}

\usepackage{xcolor}
\usepackage{graphicx}
\usepackage{enumitem}
\usepackage{multirow}
\usepackage{adjustbox}
\usepackage{booktabs}
\usepackage{pdfpages}
\usepackage{mdframed}
\usepackage{hyperref}
\usepackage{hyperref}
\usepackage{pdfpages}
\usepackage{adjustbox}
\usepackage[normalem]{ulem}
\usepackage{tabularx}
\usepackage{pdfpages}

\definecolor{customcolor}{RGB}{39, 39, 141}

\definecolor{greentext}{RGB}{100, 196, 90}
\definecolor{redtext}{RGB}{231, 73, 38}
\definecolor{bluecolor}{RGB}{39, 39, 141}
\usepackage{amsmath} 

\usepackage{graphicx}

\title{eRevise+RF: A Writing Evaluation System for Assessing Student Essay Revisions and Providing Formative Feedback}

\author{Zhexiong Liu$^{1}$, Diane Litman$^{1,2}$, Elaine Wang$^{3}$, Tianwen Li$^{2}$, Mason Gobat$^{1}$,\\ \textbf{Lindsay Clare Matsumura}$^{2}$, \textbf{Richard Correnti}$^{2}$
 \\
    $^{1}$Department of Computer Science, 
  $^{2}$Learning Research and Development Center \\
University of Pittsburgh, RAND Corporation, Pittsburgh PA USA
\\
   \texttt{zhexiong@cs.pitt.edu}, \texttt{ewang@rand.org} \\ \texttt{\{dlitman,mng28,tianwen.li,lclare,rcorrent\}@pitt.edu}
  }

\begin{document}
\maketitle
\begin{abstract}
The ability to revise essays in response to feedback is important for students' writing success. An automated writing evaluation (AWE) system that supports students in revising their essays is thus essential. We present eRevise+RF, an enhanced AWE system for assessing student essay revisions (e.g., changes made to an essay to improve its quality in response to essay feedback) and providing revision feedback. We deployed the system with 6 teachers and 406 students across 3 schools in Pennsylvania and Louisiana. The results confirmed its effectiveness in (1) assessing student essays in terms of evidence usage, (2) extracting evidence and reasoning revisions across essays, and (3) determining revision success in responding to feedback. The evaluation also suggested eRevise+RF is a helpful system for young students to improve their argumentative writing skills through revision and formative feedback.
\end{abstract}

\section{Introduction}
\label{sec:introduction}
Young student writers often struggle with identifying convincing evidence and connecting it to claims when writing argumentative essays; however, developing persuasive argumentation itself helps students improve their thinking and reasoning skills~\cite{kuhn2017argue}. This motivates recent research in developing Automated Writing Evaluation (AWE) systems for supporting students in writing and revising argumentative essays. For instance,~\citet{zhang2019erevise} developed the eRevise system to score student essays on text-based evidence usage and to provide associated feedback to guide revision. Although students attempted to respond to the feedback, their revisions often did not yield substantive essay improvement~\cite{wang2020eRevise}. Similar findings have been found for other AWE systems \cite{graham2015AWE}, suggesting that students often lack the skills necessary for effective revision \cite{10.1007/978-3-642-39112-5_27}. 

The feedback provided by~\citet{zhang2019erevise} was limited to evidence use, i.e.,\textit{``Adding more evidence would make your argument even more convincing.''} Additional feedback on revisions made in response to argumentative feedback could potentially help students self-evaluate whether their revision attempts indeed improve their essays. For instance, the feedback \textit{``When you revised your essay, it looks like you added in evidence that was very similar to the evidence you had included before. When writers revise, they generally add new content to their essays''} suggests an attempt to address feedback on evidence use was unsuccessful, because existing rather than new evidence was added. {\it We envision that using NLP to provide feedback addressing revision will further support the development of students' argumentative writing skills.}

\begin{figure}[t]
    \centering
    \includegraphics[width=0.75\linewidth]{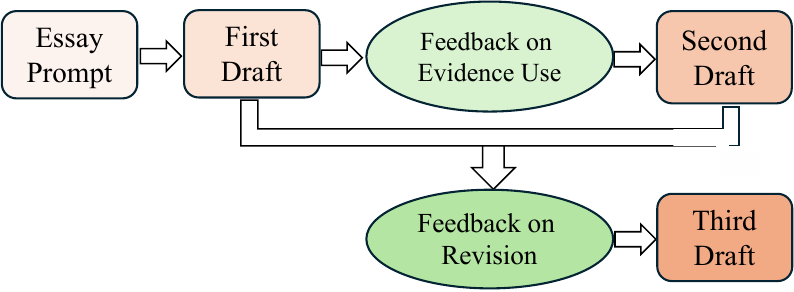}
    \caption{The eRevise+RF system usage pipeline, where students work on three essay drafts, and receive evidence use feedback and revision feedback.}
    \label{fig:introduction}
    \vspace{-1.2em}
\end{figure}

\begin{figure*}[t]
    \centering
    \includegraphics[width=0.97\linewidth]{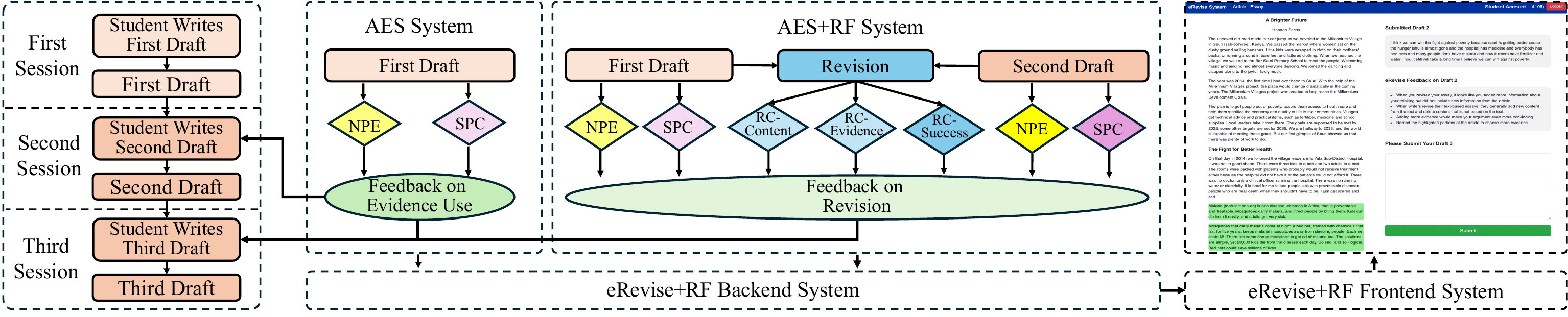}
    \vspace{-.7em}
    \caption{System architecture, including AES and AES+RF backend systems and user frontend interface. 
    The AES system generates evidence use feedback based on scoring indicators (NPE and SPC scores) and the AES+RF system provides revision feedback based on both scoring indicators and revision classifiers.
    }
    \label{fig:architecture}
    \vspace{-1em}
\end{figure*}

Motivated by this, we developed the eRevise+RF system\footnote{The eRevise+RF system is online at \url{http://erevise.lrdc.pitt.edu}. A demo video is available \href{https://youtu.be/AOyd7sVHRwg}{\textcolor{customcolor}{here}}. The source code is available at \url{https://github.com/ZhexiongLiu/eRevise-RF-System}} (RF stands for ``revision feedback'') to assess whether student revisions align with the previous feedback they receive and examine how these revisions successfully improve the argumentative essay. In particular, our system is used to support students who are taking the Response-to-Text Assessment (RTA)~\cite{correnti2013assessing}, which is an argumentative writing task for assessing students’ ability to reason using text-based evidence and to successfully use evidence to support their claims. To administer the RTA, a teacher reads a non-fiction article aloud to students as they follow along with their copy. For our pilot studies, students read an article about the United Nations Millennium Villages Project to fight poverty in Kenya (MVP) in Appendix~\ref{sec:mvp_article}, or an article about the benefits and costs of Space Exploration (SPACE) in Appendix~\ref{sec:space_article}.

Figure~\ref{fig:introduction} shows the system usage pipeline after reading the RTA article. First, students use the system to write an essay in response to a prompt, e.g., ``\textit{Based on the article, did the author provide a convincing argument that winning the fight against poverty is achievable in our lifetime? Explain why or why not with 3 to 4 examples from the text to support your answer.}'' After students submit their first essay drafts, the backend of the system automatically processes the essays by extracting evidence features (e.g., scoring indicators), which are then used to select from expert-designed messages to provide feedback on text-based evidence usage. After students receive this feedback and revise and submit second drafts of their essays, the backend system now extracts all revised sentences between the first and the second drafts and classifies whether each revision contributes to an essay's improvement in alignment with the system's prior feedback. Both the evidence features and a set of revision classifiers are used to select from expert-designed messages to provide feedback on revisions and guide students in writing their third drafts. Here, we use expert-crafted rather than LLM-generated feedback because expert feedback offers more specific and actionable suggestions to address critical issues in student essays, which is particularly helpful for young student writers. Similar findings have been confirmed in a recent study~\cite{behzad-etal-2024-assessing}.

We evaluate eRevise+RF by addressing two research questions: \textbf{RQ1}: Can the system use NLP to effectively assess student argumentative writing and revisions? \textbf{RQ2}: Can the system help students improve essays through formative feedback on their revision quality and evidence usage? 

\section{Related AWE Work on Revision}
While AWE systems typically allow submissions of multiple drafts in response to feedback, most provide feedback on single essay drafts rather than on the revisions between two drafts~\cite{wilson2021automated,huawei2023systematic,fleckenstein2023automated}. However, feedback on revisions has been identified as an important area for AWE research~\cite{guo2023automated,correnti2024supporting,li-etal-2024-using}, and some systems have visualized NLP revision analyses, e.g., to assist students in self-monitoring writing~\cite{zhang2016argrewrite,shibani2018kb,litman2022argrewrite}. While these systems display revisions, they do not provide feedback messages based on an NLP assessment of a revision's success. Although prior NLP revision work has created annotated revision corpora~\cite{zhang-etal-2017-corpus,anthonio-etal-2020-wikihowtoimprove,spangher-etal-2022-newsedits,du-etal-2022-understanding-iterative,darcy-etal-2024-aries} and has developed models for isolated tasks like identifying revision purposes~\cite{afrin2020RER,kashefi2022,du-etal-2022-understanding-iterative,mita-etal-2024-towards,jourdan-etal-2024-casimir}, generating revisions~\cite{chong-etal-2023-leveraging,ziegenbein-etal-2024-llm}, or evaluating revision quality~\cite{liu-etal-2023-predicting}, our work is the first to integrate multiple such tasks into a deployed and evaluated AWE system.

\section{System Architecture}
\subsection{eRevise+RF Backend}
The backend system (in Figure~\ref{fig:architecture}) builds on and integrates NLP algorithms from our prior research~\cite{zhang2019erevise,liu-etal-2023-predicting} to process argumentative essays, revisions between drafts, and provide feedback on both. It consists of an AES (automated essay scoring) system and an AES+RF system, of which the former provides evidence use feedback and the latter focuses on revision feedback. The AES system uses NLP algorithms to extract a set of scoring indicators and then uses the indicators to select from a set of expert-designed evidence feedback messages. The AES+RF system uses revision classifiers in addition to scoring indicators to select from another set of expert-designed revision feedback messages. 

\label{sec:evidence_score}
\textbf{Evidence Scoring Indicators.} 
The algorithms used in the AES and AES+RF systems to extract features from essays were adapted from the eRevise system~\cite{zhang2019erevise}. Given the text of an essay draft, the algorithm computes the Number of Pieces of Evidence (\textbf{NPE}), an integer encoding the number of evidence topics mentioned in the essay\footnote{The expert-crafted evidence topics include Hospital, Malaria, Farming, and School for MVP article; People, Earth, Cost, and Exploration for SPACE article~\cite{rahimi2017assessing}.}. A sliding window is applied to extract NPE on the raw text. If a window contains a similar word from a manually crafted list of keywords that are associated with evidence topics, the window is confirmed to contain text-based evidence that is related to the topic. The eight-word sliding window, Glove embedding~\cite{pennington-etal-2014-glove}, cosine similarity with a 0.9 threshold are used in the implementation based on the training data from previous RTA deployments~\cite{zhang2019erevise}. Also, the indicator Specificity (\textbf{SPC}) of the essay uses the same sliding window to determine if it contains words from another manually crafted list, which is a vector of integers that encodes the number of speciﬁc examples mentioned in 8 categories for MVP and 7 categories for SPACE, taken from \citet{rahimi2017assessing}. SPC score is the sum of the vector, whose value is a positive integer. The NPE and SPC compute the breadth and depth of the text-based evidence usage in an essay, respectively.

\begin{figure}[t]
    \centering
    \includegraphics[width=0.65\linewidth]{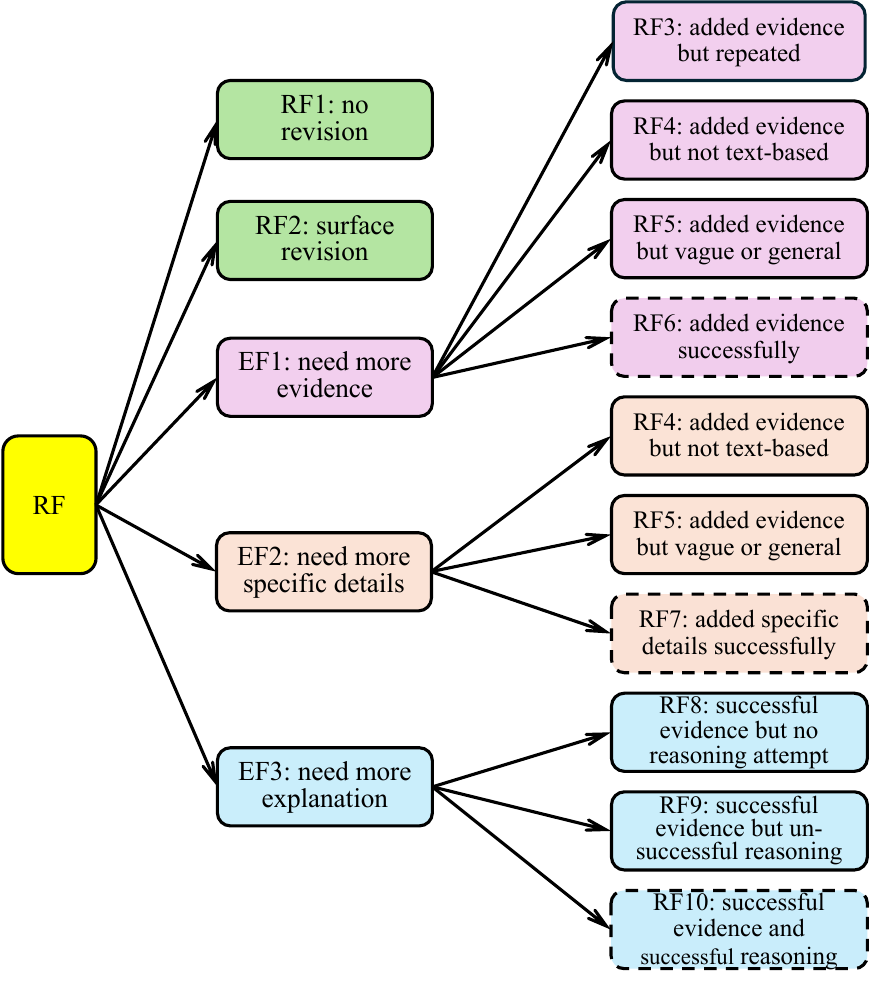}
    \vspace{-.8em}
    \caption{Revision feedback tree, where solid squares are unsuccessful revision feedback focused on helping the student try again, and the dotted squares are successful revision feedback that advances the student to a new evidence usage skill.
    }
    \label{fig:tree}
    \vspace{-1.25em}
\end{figure}

\textbf{Revision Classifiers.} \label{sec:revision_score}
In addition to the scoring indicators, the AES+RF system uses classifiers to determine if a sentence-level revision successfully contributes to an essay improvement in alignment with the feedback received. Specifically, the original and revised drafts are aligned into pairs of original and revised sentences using the sentence alignment tool Bertalign~\cite{liu2022bertalign}. The pairs of non-identically aligned sentences are extracted as the essay revisions. A BERT-based~\cite{devlin-etal-2019-bert} binary revision purpose classifier (\textbf{RC-Content}) that is trained on a college-level revision corpus~\cite{kashefi2022} is used to classify essay revisions into surface (meaning-preserving) and content (meaning-altering) revisions. Note that the surface revisions are later removed because the system is only interested in content revisions. Another binary revision classifier (\textbf{RC-Evidence}) is used to predict whether each content revision is an evidence or reasoning revision, motivated by the prior work that assesses revision quality from the perspective of evidence and reasoning~\cite{afrin2020RER}. RC-Evidence is implemented with ChatGPT, which returns an evidence or reasoning label given a revision pair and the prompt ``\textit{You need to identify whether the given sentence is an evidence or reasoning sentence. Your output should be chosen from the list [evidence, reasoning]}''. Furthermore, a binary revision classifier (\textbf{RC-Success}) predicts whether each content revision is successful or unsuccessful. RC-Success is a DistilRoBERTa model~\cite{sanh2019distilbert} developed based on~\citet{liu-etal-2023-predicting}. It leverages Argument Context (AC) in addition to revision pairs to determine successful vs. unsuccessful revisions. The ACs are extracted from essays with Chain-of-Thought prompts for ChatGPT used in prior work~\cite{liu-etal-2023-predicting}. Finally, the combination of RC-Evidence and RC-Success yields 4 revision labels, i.e., successful evidence, unsuccessful evidence, successful reasoning, and unsuccessful reasoning. The model implementation details are described in Appendix~\ref{appendix:implementation}.

\begin{table*}[t]
\centering
\tiny
\begin{tabular}{c|ccccccc|ccccccc}
\hline
\multirow{3}{*}{Student} & \multicolumn{7}{c|}{MVP}                                                                                                     & \multicolumn{7}{c}{SPACE}                                                                                                   \\ \cline{2-15} 
                       & \multicolumn{2}{c|}{Essay}     & \multicolumn{3}{c|}{Revision Action}                   & \multicolumn{2}{c|}{Revision Type} & \multicolumn{2}{c|}{Essay}     & \multicolumn{3}{c|}{Revision Action}                   & \multicolumn{2}{c}{Revision Type} \\ \cline{2-15} 
                       & N   & \multicolumn{1}{c|}{WC}  & Add       & Delete    & \multicolumn{1}{c|}{Modify}    & Surface         & Content          & N   & \multicolumn{1}{c|}{WC}  & Add       & Delete    & \multicolumn{1}{c|}{Modify}    & Surface         & Content         \\ \hline
Grade 4                      & 165 & \multicolumn{1}{c|}{227} & 4.0 / 4.5 & 1.0 / 1.9 & \multicolumn{1}{c|}{2.3 / 2.1} & 2.2 / 1.8       & 5.1 / 6.7        & 72  & \multicolumn{1}{c|}{176} & 5.3 / 5.2 & 2.4 / 3.2 & \multicolumn{1}{c|}{1.9 / 2.1} & 1.7 / 1.8       & 7.8 / 8.7       \\
Grade 5                      & 72  & \multicolumn{1}{c|}{323} & 9.3 / 5.5 & 6.8 / 2.4 & \multicolumn{1}{c|}{3.5 / 3.3} & 2.5 / 2.9       & 17.1 / 8.4       & 129 & \multicolumn{1}{c|}{248} & 5.9 / 3.9 & 2.0 / 1.8 & \multicolumn{1}{c|}{1.8 / 2.6} & 1.6 / 2.3       & 8.1 / 6.0       \\
Grade 6                      & 18  & \multicolumn{1}{c|}{324} & 8.0 / 7.8 & 2.8 / 7.7 & \multicolumn{1}{c|}{2.3 / 3.0} & 2.8 / 2.2       & 10.3 / 16.3      & -   & \multicolumn{1}{c|}{-}   & -         & -         & \multicolumn{1}{c|}{-}         & -               & -               \\
Grade 7                      & 222 & \multicolumn{1}{c|}{374} & 4.4 / 2.1 & 1.6 / 0.7 & \multicolumn{1}{c|}{2.0 / 1.1} & 1.8 / 1.1       & 6.2 / 2.7        & 225 & \multicolumn{1}{c|}{337} & 4.5 / 3.3 & 1.6 / 1.6 & \multicolumn{1}{c|}{2.6 / 2.0} & 2.5 / 1.8       & 6.2 / 5.0       \\
Grade 8                      & 39  & \multicolumn{1}{c|}{399} & 3.5 / 2.5 & 1.5 / 0.9 & \multicolumn{1}{c|}{1.1 / 1.8} & 1.5 / 1.4       & 4.5 / 3.8        & 24  & \multicolumn{1}{c|}{363} & 3.1 / 6.8 & 1.4 / 8.3 & \multicolumn{1}{c|}{2.1 / 2.8} & 2.1 / 2.6       & 4.5 / 15.1      \\\hline
Overall                      & 516  & \multicolumn{1}{c|}{320} &  5.0 / 3.6  &  2.2 / 1.6  & \multicolumn{1}{c|}{2.2 / 1.8} & 2.0 / 1.7       & 7.4 / 5.4        & 450  & \multicolumn{1}{c|}{287} &  5.0 / 3.9  & 1.8 / 2.3 & \multicolumn{1}{c|}{2.2 / 2.2} & 2.1 / 2.0       & 6.9 / 6.4      \\\hline
\end{tabular}
\vspace{-1.3em}
\caption{The statistics of the collected MVP and SPACE essays in terms of grades, number of essays (N), average word counts (WC) in an essay, and average numbers of revision actions and revision types between draft1-draft2 (before the slash) and draft2-draft3 (after the slash), respectively.}
\label{tab:essay-stats}
\vspace{-1em}
\end{table*}

\textbf{Evidence Use Feedback.}
The AES system provides 3 levels of Evidence Use Feedback (EF). EF1 feedback focuses solely on providing more evidence (completeness); EF2 feedback focuses on providing more details (specificity); and EF3 feedback focuses on explaining and connecting evidence to an argument (explanation). The AES system uses scoring indicators (NPE and SPC) with expert-designed thresholds ($\alpha$ and $\beta$) to determine which feedback level message to display after essay submission. The detailed algorithm and feedback messages are described in Appendix~\ref{sec:evidence_use_feedback_selection}.

\textbf{Revision Feedback.} The AES+RF system provides 10 levels of Revision Feedback (RF) organized on a revision feedback tree in Figure~\ref{fig:tree}, after assessing whether a student successfully follows the EF to revise the last essay draft. Thus RF focuses on the revision process rather than on the essay content itself. For a student who attempts to revise based on EF previously received, either unsuccessful revision feedback is provided to help the student try again, or successful revision feedback is used to help advance the student's essay into the next level. The system uses both scoring indicators and revision classifiers to determine the RF. The detailed algorithm and feedback messages are described in Appendix~\ref{sec:revision_feedback_selection}.

\subsection{eRevise+RF Frontend}
eRevise+RF is a web-based system that offers three types of interfaces for students, teachers, and administrators. The student interface (Figure~\ref{fig:interface-student} in the Appendix; also see Figure~\ref{fig:architecture}) displays the student's last submitted essay along with its feedback messages. It also allows the student to submit a revised draft. The teacher interface (Figure~\ref{fig:interface-teacher} in the Appendix) includes a submission page that monitors all students' submitted drafts and feedback messages. It also has a classroom page for retrieving student information. The administrator interface has the highest privilege of accessing teacher and student interfaces as well as managing users (e.g., adding and deleting users).

\section{System Deployment}
The eRevise+RF system was deployed in collaboration with 6 teachers and 406 students in grades 4 to 8 during Spring 2024 for the MVP article and Fall 2024 for the SPACE article. The teachers overlapped for the 2 deployments, while the students were different (e.g., no grade 6 students in Fall). The participants were from one school in Pennsylvania (PA), where 35\% of the students are students of color, representing 45 different zip code areas, and speaking more than 10 languages at home; two schools in Louisiana (LA), where 98\% and 65\% enrollment are minorities, respectively, and student reading proficiency rates range from 46\% to 47\%. 

The system was deployed in three sessions administered over 3 to 5 days (see Figure~\ref{fig:architecture}). The 2 deployments received a total of 194 draft1, 194 draft2, and 172 draft3 for MVP and 176 draft1, 176 draft2, and 150 draft3 for SPACE essays since students might not have consistently participated in all three sessions due to absences. Our analysis is based on the essays from students who participated in all three sessions, yielding 3 drafts of 172  MVP essays (516 total) and of 150  SPACE essays (450 total). Table~\ref{tab:essay-stats} shows the essay revision statistics across grades, where higher-grade students usually have longer essays, and students are inclined to make adding revisions across all grades. 

\section{Results and Analysis}
\begin{table}[t]
\tiny
\centering
\begin{tabular}{ccc|ccc}
\hline
         & \multicolumn{2}{c|}{\begin{tabular}[c]{@{}c@{}}Evidence Scoring \\ Indicator (QWK)\end{tabular}} & \multicolumn{3}{c}{\begin{tabular}[c]{@{}c@{}}Revision Classifier \\ (F1)\end{tabular}} \\ \cline{2-6} 
         & NPE                                             & SPC                                            & RC-Con                           & RC-Evi                           & RC-Suc                           \\ \hline
Values    & 0.67                                            & 0.82                                           & 0.96                           & 0.66                           & 0.70                           \\
Size (N) & 516                                             & 516                                            & 1,525                            & 1,046                          & 1,024                          \\ \hline
\end{tabular}

\vspace{-1.5em}
\caption{The evaluation of essay-level evidence scoring indicators and sentence-level revision classifiers based on predicted vs. human annotations on MVP essays.}
\label{tab:kappa-npe-spc}
\vspace{-2.55em}
\end{table}

\subsection{System NLP Evaluation}
\label{sec:analysis}
We use the 516 MVP essays for manual annotation, to evaluate the accuracy of the scoring indicator algorithms. We employed experts to annotate NPE and SPC scores for every sampled essay. The expert annotators participated in prior studies with Kappas of 0.8 and 0.84 for Evidence and Reasoning~\cite{liu-etal-2023-predicting}, and ICCs of 0.84 and 0.69 for NPE and SPC~\cite{CORRENTI2022100084}, respectively. Table~\ref{tab:kappa-npe-spc} shows that Quadratic Weighted Kappa (QWK) between system-predicted and human-annotated NPE and SPC are 0.67 and 0.82, respectively. The agreements are substantial and almost perfect, respectively~\cite{landis77}, which suggests the scoring indicators (NPE and SPC) are effective. Also, we investigate the performance of the EF level predictions on the 172 first drafts since this EF is mainly designed for the first drafts. The QWK between the predicted and annotated evidence feedback levels\footnote{We feed expert-annotated NPE and SPC to level selection algorithm (in Appendix~\ref{sec:evidence_use_feedback_selection}) to obtain annotated EF levels.} is 0.66, which suggests the AES(+RF) system used to select evidence use feedback levels for students' first drafts is also reasonably effective.

\begin{table}[t]
\centering
\tiny
\begin{tabular}{ccccc|cccc}
\hline
\multirow{2}{*}{Essay} & \multicolumn{4}{c|}{MVP  (N=516)} & \multicolumn{4}{c}{SPACE (N=450)} \\ \cline{2-9} 
                       & WC    & ES     & NPE    & SPC    & WC     & ES     & NPE    & SPC    \\ \hline
Draft 1                      & 293   & 2.45   & 2.56   & 8.81   & 228    & 2.09   & 3.02   & 6.07   \\
Draft 2                      & 361   & 2.81   & 3.19   & 12.21  & 295    & 2.55   & 3.84   & 7.88   \\
Draft 3                      & 405   & 2.91   & 3.34   & 13.98  & 338    & 2.85   & 4.09   & 8.84   \\ \hline
\end{tabular}

\vspace{-1.5em}
\caption{Essay quality improvement based on WC and predicted values of  ES, NPE, and SPC metrics.}
\label{tab:improvment}
    \vspace{-2.2em}
\end{table}

\begin{table*}[t]
\centering
\tiny
\begin{tabular}{cc|cccc|cccc}
\hline
\multicolumn{2}{c|}{\multirow{2}{*}{\begin{tabular}[c]{@{}c@{}}Evidence \\ Feedback\end{tabular}}}             & \multicolumn{4}{c|}{Draft1-Draft2}                                         & \multicolumn{4}{c}{Draft2-Draft3}                                                       \\ \cline{3-10} 
\multicolumn{2}{c|}{}                                                                                          & NPE         & \multicolumn{1}{c|}{$\Delta$NPE} & SPC         & $\Delta$SPC & NPE                      & \multicolumn{1}{c|}{$\Delta$NPE} & SPC         & $\Delta$SPC \\ \hline
\multicolumn{1}{c|}{\multirow{3}{*}{\begin{tabular}[c]{@{}c@{}}Annotated\\ NPE/SPC\end{tabular}}} & EF1 (N=74) & 1.09 / 2.49 & \multicolumn{1}{c|}{+128\%}     & 4.41 / 9.18 & +108\%     & 2.49 / 2.77              & \multicolumn{1}{c|}{+11.2\%}     & 9.18 / 11.7 & +27.8\%     \\
\multicolumn{1}{c|}{}                                                                             & EF2 (N=11) & 2.64 / 3.18 & \multicolumn{1}{c|}{+20.5\%}     & 6.45 / 10.0 & +55.0\%     & 3.18 / 2.82              & \multicolumn{1}{c|}{-11.3\%}     & 10.0 / 10.5 & +4.50\%     \\
\multicolumn{1}{c|}{}                                                                             & EF3 (N=87) & 3.03 / 3.09 & \multicolumn{1}{c|}{+1.98\%}     & 12.9 / 14.6 & +13.0\%     & 3.09 / 3.08 & \multicolumn{1}{c|}{-0.00\%}     & 14.6 / 15.6 & +6.50\%     \\ \hline
\multicolumn{1}{c|}{\multirow{3}{*}{\begin{tabular}[c]{@{}c@{}}Predicted\\ NPE/SPC\end{tabular}}} & EF1 (N=74) & 1.28 / 2.74 & \multicolumn{1}{c|}{+114\%}      & 4.78 / 10.1 & +112\%      & 2.74 / 3.05              & \multicolumn{1}{c|}{+11.3\%}     & 10.1 / 12.7 & +24.9\%     \\
\multicolumn{1}{c|}{}                                                                             & EF2 (N=11) & 3.18 / 3.45 & \multicolumn{1}{c|}{+8.49\%}     & 3.09 / 6.00 & +94.2\%     & 3.45 / 3.09              & \multicolumn{1}{c|}{-10.4\%}     & 6.00 / 7.73 & +28.8\%     \\
\multicolumn{1}{c|}{}                                                                             & EF3 (N=87) & 3.56 / 3.53 & \multicolumn{1}{c|}{-0.84\%}     & 13.0 / 14.7 & +13.4\%     & 3.53 / 3.61              & \multicolumn{1}{c|}{+2.27\%}     & 14.7 / 15.9 & +7.75\%     \\ \hline
\end{tabular}

\vspace{-1.4em}
\caption{Annotated and predicted NPE/SPC changes between two MVP essay drafts with respect to received evidence feedback levels on the old drafts. The number before the slash is the NPE and SPC on the old drafts and the latter is on the new drafts. $\Delta$ means the ratio of the changes from the old drafts to the new drafts. Both evidence feedback (from draft1 to draft2) and revision feedback (from draft2 to draft3) help students improve their essays.
}
\label{tab:change-npe-spc}
    \vspace{-1.75em}
\end{table*}

Again, we employed the trained annotators to annotate 172 pairs of first and second MVP essay drafts, to evaluate the performance of the revision classifiers. The annotators annotated \textit{surface} or \textit{content} for 1,525 revision pairs, and \textit{evidence} or \textit{reasoning} labels for 1,046 out of 1,525 extracted revision pairs, given the other 479 revision pairs were labeled with \textit{surface}, and were removed since RC-Evidence only handles content revisions. Moreover, the annotators annotated the RER scheme~\cite{afrin2020RER} on the revision pairs based on the prior work~\cite{liu-etal-2023-predicting} and obtained 1,024 annotated \textit{successful} and \textit{unsuccessful} labels, given the other 22 revision pairs were labeled with \textit{claim}, which were removed since RC-Success only handles evidence and reasoning revisions. Table~\ref{table:data-example-alignment} in the Appendix shows an annotated essay example. 

Using these annotations to evaluate the three revision classifiers yields RC-Content, RC-Evidence, and RC-Success F1 scores of 0.96, 0.66, and 0.70, respectively (shown in Table~\ref{tab:kappa-npe-spc}). Precision/recall figures and a grade breakdown are shown in Table~\ref{tab:f1-recall-precision-classifier} in the Appendix, and suggest that RC-Evidence has relatively lower F1 in grade 8 while RC-Success has relatively lower precision in grades 6 and 7.  Confusion matrices are shown in Table~\ref{tab:confusion-classifier} in the Appendix,  and suggest that RC-Evidence is likely to predict evidence as reasoning and RC-Success makes errors mostly on true unsuccessful revisions. In sum, both the evidence score indicators and revision classifiers are reasonably effective in using NLP to assess student argumentative writing and revisions (RQ1).

\subsection{Student Performance} 
\label{sec:student_performance}
Similar to prior RTA studies \cite{zhang2019erevise,CORRENTI2022100084}, we evaluated changes in essay quality across drafts using the metrics of Word Count (WC), Evidence Score (ES), NPE and SPC. The NPE and SPC are predicted based on the algorithms described earlier in the evidence score indicators (in Sec.~\ref{sec:evidence_score}). The ES is predicted with a BERT~\cite{devlin-etal-2019-bert} model trained on a previously collected corpus with 0.89 QWK on MVP essays and 0.94 QWK on SPACE essays\footnote{The corpus annotated evidence scores on 317 MVP and 109 SPACE essays using a scale of 1 to 4 (low to high). }. Table~\ref{tab:improvment} shows that essay quality improves over three drafts, where for all metrics, higher is better. Specifically, MVP essays have a 24.6\% NPE increase and a 38.5\% SPC increase, while SPACE essays have a 27.2\% NPE increase and a 29.8\% SPC increase from the first to the second drafts, respectively. ES increases over the three drafts, with the greatest increase (14.69\% on MVP and 22.0\% on SPACE) occurring from the first to the second drafts. These results reveal observable improvement of the essays from the first to the second draft after evidence use feedback, and from the second to the third drafts after revision feedback.

Additionally, we use the same manually annotated 516 student essays as in Table~\ref{tab:kappa-npe-spc} to evaluate the quality changes between drafts, now using gold rather than predicted values. The annotated NPE improves on average from 2.20, 2.79, to 2.93 in the first, second, and third drafts, respectively, while the predicted NPE improves from 2.56, 3.19, to 3.34 (shown in Table~\ref{tab:improvment}). Although the system tends to overscore NPE, the inferences with respect to improvement across drafts are still the same (e.g.,~26.8\% vs. 24.6\% improvement from the first to the second drafts). The annotated SPC improves from 9.00, 12.04, to 13.59 across drafts, while the predicted SPC improves from 8.81, 12.21, to 13.98 (shown in Table~\ref{tab:improvment}), which suggests using predictions yields the same conclusions as annotations regarding improvement in essay quality. 

\subsection{Feedback Effectiveness}
\label{sec:feedback_effectiveness}
To better understand whether students are responding to the particular feedback message they receive,  we investigate both annotated and predicted NPE and SPC changes between two MVP drafts (old drafts before and new drafts after receiving feedback) in Table~\ref{tab:change-npe-spc}. For the draft1-draft2 essays, the annotated NPE increases 128\% when the old drafts receive EF1 (needing more evidence) feedback, which suggests that students do follow the EF1 to add more evidence, given NPE measures the number of pieces of evidence used. Also, the annotated SPC increases 55.0\% when the old drafts receive EF2 (needing more specific details), which also confirms students' essays have an improvement in specificity after revising essays based on EF2. When the old drafts had EF3 (needing more explanation), annotated NPE mostly remains no change ($\Delta$NPE=+1.98\%) and SPC has slight improvement. This is because the RF guided students to add explanations, not evidence. These observations confirm that how student revisions change their evidence usage is as expected for all three EF levels. Similarly, we compare draft2-draft3 essays before and after receiving revision feedback. We observe annotated NPE improvement when draft2 has EF1 (e.g., RF was either RF3, RF4, RF5, or RF6 in Figure~\ref{fig:tree}), which suggests revision feedback helps students add more evidence. However, revision feedback may not be helpful when draft2 receives EF2 given the annotated NPE average drops by 11.3\%, which suggests that student essays might slightly weaken the completeness of the evidence while revising toward specificity. These observations are consistent when using predicted NPE and SPC in Table~\ref{tab:change-npe-spc}. Also, we analyze predicted NPE and SPC changes for SPACE essays in Table~\ref{tab:change-npe-spc-SPACE} in the Appendix, which shows similar patterns. In sum, the average increases of NPE and SPC for two RTA essays suggest that both evidence use feedback and revision feedback are helpful for students to improve their writing, in ways that are in alignment with the system's feedback messages (RQ2).

\subsection{Case Study}
\begin{figure}[t]
    \centering
    \includegraphics[width=0.98\linewidth]{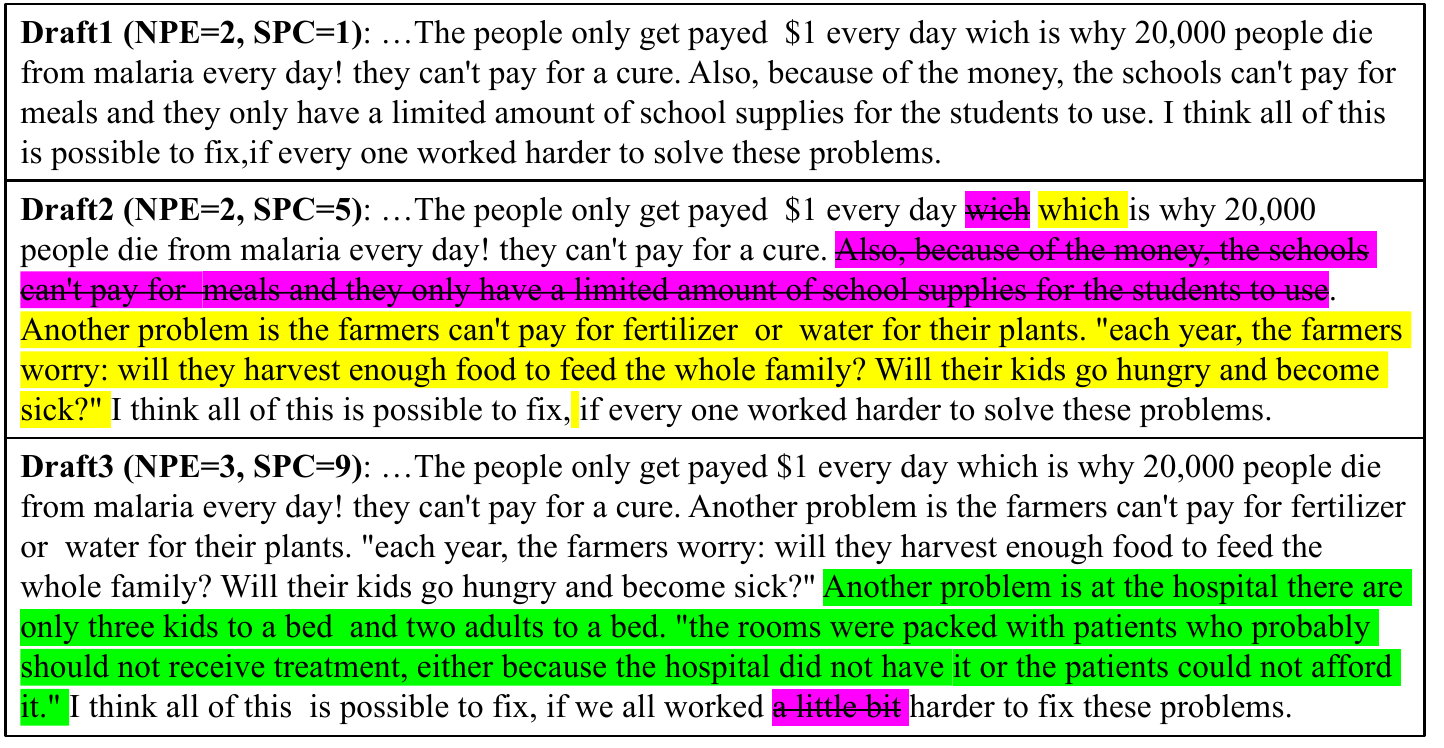}
    \vspace{-0.7em}
    \caption{The example of a MVP essay, revising from draft1 to draft2, then from draft2 to draft3. The purple marks deletion, yellow and green marks addition.}
    \label{fig:example-1}
    \vspace{-1.5em}
\end{figure}

We show a student's three argumentative essay drafts in Figure~\ref{fig:example-1} to illustrate how the feedback helps the student improve the essay through revision. The first draft does not include sufficient evidence to support the claim as it has predicted NPE of 2. Thus, the student receives EF that ``\textit{adding more evidence would make your argument even more convincing.}'' Afterward, the student makes surface (typo) and content revisions in the second draft by deleting one piece of evidence about \textit{school} (purple) and adding another piece of evidence about \textit{farming} (yellow). The second draft has the same predicted NPE of 2 but SPC increases from 1 to 5 since it replaces \textit{school} with \textit{farming} and adds more details about \textit{farming}. Additional RF is provided ``\textit{...When writers revise their text-based essays, they generally add new content from the text and delete content that is not based on the text.}'' In response, the student makes additional surface and content revisions in the third draft by adding a piece of evidence about \textit{hospital} with more details, which increases predicted NPE to 3 and SPC to 9. This example suggests the feedback is helpful for the student to reflect on revisions that have been made and further revise the essay. We show additional cases in Table~\ref{fig:example-2} and~\ref{fig:example-3} in the Appendix.

\section{Conclusion}
We present eRevise+RF, an advanced version of the eRevise system that extends and integrates algorithms to assess argumentative essay revisions, and to provide formative feedback on both evidence usage and revision success in response to feedback. Although the algorithms are primarily adapted from our prior published work, their integration into a complete and deployed system is a novel contribution. Through our pilot studies, we show promise of using NLP algorithms to scaffold young students in writing and revising text-based argumentative essays. In future work, we will deploy the system in more classrooms and with other RTA articles, to generalize our findings.

\section*{Ethics}
eRevise+RF was used to collect our essay corpus under standard protocols approved by an institutional review board (IRB). Our data is not publicly available at the moment to ensure the safety of the private information of young students. The system is currently accessible only to the teachers and students participating in our study who have been assigned user accounts, in compliance with IRB constraints and data privacy requirements that restrict open user registration. However, the system source code is made publicly available on GitHub to support contributions to the NLP community. The system does not pose any ethical concerns because of the limited access to the data, but there might be a risk that the system may give poor advice based on incorrect scoring indicators or classifier predictions, or that the predictive models may learn biases due to small annotated training corpora. However, due to restrictions of IRB and policies of partner schools, only school-level but not student-level demographics are available which makes it difficult to evaluate potential biases.

\section*{Limitations}
\textbf{System.} Although the eRevise+RF system demonstrates promise in helping students revise essays in response to formative feedback, its evaluation is based on a relatively small essay corpus (i.e., 516 MVP and 450 SPACE essays). Nonetheless, the corpus exhibits substantial diversity in both grade levels and demographics. For example, one school in PA includes students from 45 different zip code areas, while two schools in LA have minority enrollments of 98\% and 65\%, respectively. The analyses in Sec.~\ref{sec:student_performance} demonstrate the robustness of students’ essay improvement across these diverse settings. Although the system deployments were pilot and lacked a control group where students wrote without receiving feedback since we didn't have enough classrooms to make this feasible, it is a goal of future studies. However, our analysis in Sec.~\ref{sec:feedback_effectiveness} was designed to start teasing apart the impact of the system on students compared to unscaffolded revision. Our results show that the student revisions were in fact largely responsive to the feedback. Although we currently only assess the evidence dimension of the RTA, we started there because a persistent criticism of AWE systems is that they have not been designed to meet ambitious writing standards such as text-based argument writing~\cite{burstein2020expanding}. But, we note other aspects of writing quality such as creativity need to be studied in future work.

\textbf{LLM.} Our methods are mostly based on rubrics designed by RTA experts rather than advanced LLM. This is because we notice that for our deployment with young students, hallucinations and other LLM generation concerns such as offensive content and biases would need to be carefully studied and addressed. Additionally, a recent study~\cite{behzad-etal-2024-assessing} compares LLM-generated essay feedback with human feedback and concludes that human feedback is more effective in delivering specific and actionable comments that address the most critical issues in an essay. This suggests young student learners may benefit more from expert feedback tailored to their specific writing challenges. In future work, we aim to integrate both expert and LLM-generated feedback into the system to provide nuanced and adaptive feedback messages.

\textbf{Algorithm.} The evidence scoring indicators described in Sec~\ref{sec:evidence_score} sometimes overcount because the text-based evidence is not perfectly captured. Since the initial deployment, the collected and annotated essays have been used to tune the sliding window and text similarity algorithms in a data-driven manner. We have since modified the NPE algorithm based on an error analysis and it has now achieved a QWK of 0.87 compared to the original 0.67 in Table~\ref{tab:kappa-npe-spc}. The scoring indicator's reliance on manual keywords may not adapt well to different writing topics or styles unless continuously updated. Thus, other methods~\cite{zhang-litman-2020-automated,zhang-litman-2021-essay} to avoid dependence on manual keyword identification could be implemented. Also, the trained revision classifiers RC-Evidence and RC-Success are not excellent, thus may cause the AES+RF system to provide incorrect revision feedback, which could be improved in future work. 

\bibliography{custom}

\begin{thebibliography}{38}
\providecommand{\natexlab}[1]{#1}

\bibitem[{Afrin et~al.(2020)Afrin, Wang, Litman, Matsumura, and
  Correnti}]{afrin2020RER}
Tazin Afrin, Elaine~Lin Wang, Diane Litman, Lindsay~Clare Matsumura, and
  Richard Correnti. 2020.
\newblock Annotation and classification of evidence and reasoning revisions in
  argumentative writing.
\newblock In \emph{Proceedings of the Fifteenth Workshop on Innovative Use of
  NLP for Building Educational Applications}, Seattle, Washington, USA
  (Remote).

\bibitem[{Anthonio et~al.(2020)Anthonio, Bhat, and
  Roth}]{anthonio-etal-2020-wikihowtoimprove}
Talita Anthonio, Irshad Bhat, and Michael Roth. 2020.
\newblock \href {https://aclanthology.org/2020.lrec-1.702}
  {wiki{H}ow{T}o{I}mprove: A resource and analyses on edits in instructional
  texts}.
\newblock In \emph{Proceedings of the Twelfth Language Resources and Evaluation
  Conference}, pages 5721--5729, Marseille, France. European Language Resources
  Association.

\bibitem[{Behzad et~al.(2024)Behzad, Kashefi, and
  Somasundaran}]{behzad-etal-2024-assessing}
Shabnam Behzad, Omid Kashefi, and Swapna Somasundaran. 2024.
\newblock \href {https://aclanthology.org/2024.lrec-main.144} {Assessing online
  writing feedback resources: Generative {AI} vs. good samaritans}.
\newblock In \emph{Proceedings of the 2024 Joint International Conference on
  Computational Linguistics, Language Resources and Evaluation (LREC-COLING
  2024)}, pages 1638--1644, Torino, Italia. ELRA and ICCL.

\bibitem[{Burstein et~al.(2020)Burstein, Riordan, and
  McCaffrey}]{burstein2020expanding}
Jill Burstein, Brian Riordan, and Daniel McCaffrey. 2020.
\newblock Expanding automated writing evaluation.
\newblock In \emph{Handbook of automated scoring}, pages 329--346. Chapman and
  Hall/CRC.

\bibitem[{Chong et~al.(2023)Chong, Kong, Wu, Liu, Jin, Yang, Fan, Fan, and
  Yang}]{chong-etal-2023-leveraging}
Ruining Chong, Cunliang Kong, Liu Wu, Zhenghao Liu, Ziye Jin, Liner Yang, Yange
  Fan, Hanghang Fan, and Erhong Yang. 2023.
\newblock \href {https://doi.org/10.18653/v1/2023.acl-short.105} {Leveraging
  prefix transfer for multi-intent text revision}.
\newblock In \emph{Proceedings of the 61st Annual Meeting of the Association
  for Computational Linguistics (Volume 2: Short Papers)}, pages 1219--1228,
  Toronto, Canada. Association for Computational Linguistics.

\bibitem[{Correnti et~al.(2013)Correnti, Matsumura, Hamilton, and
  Wang}]{correnti2013assessing}
Richard Correnti, Lindsay~Clare Matsumura, Laura Hamilton, and Elaine Wang.
  2013.
\newblock Assessing students' skills at writing analytically in response to
  texts.
\newblock \emph{The Elementary School Journal}, 114(2):142--177.

\bibitem[{Correnti et~al.(2022)Correnti, Matsumura, Wang, Litman, and
  Zhang}]{CORRENTI2022100084}
Richard Correnti, Lindsay~Clare Matsumura, Elaine~Lin Wang, Diane Litman, and
  Haoran Zhang. 2022.
\newblock \href {https://doi.org/10.1016/j.caeo.2022.100084} {Building a
  validity argument for an automated writing evaluation system (erevise) as a
  formative assessment}.
\newblock \emph{Computers and Education Open}, 3:100084.

\bibitem[{Correnti et~al.(2024)Correnti, Wang, Matsumura, Litman, Liu, and
  Li}]{correnti2024supporting}
Rip Correnti, Elaine~Lin Wang, Lindsay~Clare Matsumura, Diane Litman, Zhexiong
  Liu, and Tianwen Li. 2024.
\newblock Supporting students' text-based evidence use via formative automated
  writing and revision assessment.
\newblock In \emph{The Routledge international handbook of automated essay
  evaluation}, pages 221--243. Routledge.

\bibitem[{D{'}Arcy et~al.(2024)D{'}Arcy, Ross, Bransom, Kuehl, Bragg, Hope, and
  Downey}]{darcy-etal-2024-aries}
Mike D{'}Arcy, Alexis Ross, Erin Bransom, Bailey Kuehl, Jonathan Bragg, Tom
  Hope, and Doug Downey. 2024.
\newblock \href {https://doi.org/10.18653/v1/2024.acl-long.377} {{ARIES}: A
  corpus of scientific paper edits made in response to peer reviews}.
\newblock In \emph{Proceedings of the 62nd Annual Meeting of the Association
  for Computational Linguistics (Volume 1: Long Papers)}, pages 6985--7001,
  Bangkok, Thailand. Association for Computational Linguistics.

\bibitem[{Devlin et~al.(2019)Devlin, Chang, Lee, and
  Toutanova}]{devlin-etal-2019-bert}
Jacob Devlin, Ming-Wei Chang, Kenton Lee, and Kristina Toutanova. 2019.
\newblock \href {https://doi.org/10.18653/v1/N19-1423} {{BERT}: Pre-training of
  deep bidirectional transformers for language understanding}.
\newblock In \emph{Proceedings of the 2019 Conference of the North {A}merican
  Chapter of the Association for Computational Linguistics: Human Language
  Technologies, Volume 1 (Long and Short Papers)}, pages 4171--4186,
  Minneapolis, Minnesota. Association for Computational Linguistics.

\bibitem[{Du et~al.(2022)Du, Raheja, Kumar, Kim, Lopez, and
  Kang}]{du-etal-2022-understanding-iterative}
Wanyu Du, Vipul Raheja, Dhruv Kumar, Zae~Myung Kim, Melissa Lopez, and Dongyeop
  Kang. 2022.
\newblock \href {https://doi.org/10.18653/v1/2022.acl-long.250} {Understanding
  iterative revision from human-written text}.
\newblock In \emph{Proceedings of the 60th Annual Meeting of the Association
  for Computational Linguistics (Volume 1: Long Papers)}, pages 3573--3590,
  Dublin, Ireland. Association for Computational Linguistics.

\bibitem[{Fleckenstein et~al.(2023)Fleckenstein, Liebenow, and
  Meyer}]{fleckenstein2023automated}
Johanna Fleckenstein, Lucas~W Liebenow, and Jennifer Meyer. 2023.
\newblock Automated feedback and writing: A multi-level meta-analysis of
  effects on students' performance.
\newblock \emph{Frontiers in Artificial Intelligence}, 6:1162454.

\bibitem[{Graham et~al.(2015)Graham, Harris, and Santangelo}]{graham2015AWE}
Steve Graham, Karen~R Harris, and Tanya Santangelo. 2015.
\newblock Research-based writing practices and the common core: Meta-analysis
  and meta-synthesis.
\newblock \emph{The Elementary School Journal}, 115(4):498--522.

\bibitem[{Guo et~al.(2023)Guo, Feng, and Hua}]{guo2023automated}
Qian Guo, Ruiling Feng, and Yuanfang Hua. 2023.
\newblock \emph{Automated written corrective feedback in Research Paper
  Revision: The good, the bad, and the missing}.
\newblock Routledge.

\bibitem[{Huawei and Aryadoust(2023)}]{huawei2023systematic}
Shi Huawei and Vahid Aryadoust. 2023.
\newblock A systematic review of automated writing evaluation systems.
\newblock \emph{Education and Information Technologies}, 28(1):771--795.

\bibitem[{Jourdan et~al.(2024)Jourdan, Boudin, Hernandez, and
  Dufour}]{jourdan-etal-2024-casimir}
L{\'e}ane Jourdan, Florian Boudin, Nicolas Hernandez, and Richard Dufour. 2024.
\newblock \href {https://aclanthology.org/2024.lrec-main.257} {{CASIMIR}: A
  corpus of scientific articles enhanced with multiple author-integrated
  revisions}.
\newblock In \emph{Proceedings of the 2024 Joint International Conference on
  Computational Linguistics, Language Resources and Evaluation (LREC-COLING
  2024)}, pages 2883--2892, Torino, Italia. ELRA and ICCL.

\bibitem[{Kashefi et~al.(2022)Kashefi, Afrin, Dale, Olshefski, Godley, Litman,
  and Hwa}]{kashefi2022}
Omid Kashefi, Tazin Afrin, Meghan Dale, Christopher Olshefski, Amanda Godley,
  Diane Litman, and Rebecca Hwa. 2022.
\newblock \href {https://doi.org/10.1007/s10579-021-09567-z} {Argrewrite v.2:
  an annotated argumentative revisions corpus}.
\newblock \emph{Language Resources and Evaluation}, pages 1574--0218.

\bibitem[{Kuhn et~al.(2017)Kuhn, Hemberger, and Khait}]{kuhn2017argue}
Deanna Kuhn, Laura Hemberger, and Valerie Khait. 2017.
\newblock \emph{Argue with me: Argument as a path to developing students'
  thinking and writing}.
\newblock Routledge.

\bibitem[{Landis and Koch(1977)}]{landis77}
J.~Richard Landis and Gary Koch. 1977.
\newblock The measurement of observer agreement for categorical data.
\newblock \emph{Biometrics}, 33(1):159--174.

\bibitem[{Li et~al.(2024)Li, Liu, Matsumura, Wang, Litman, and
  Correnti}]{li-etal-2024-using}
Tianwen Li, Zhexiong Liu, Lindsay Matsumura, Elaine Wang, Diane Litman, and
  Richard Correnti. 2024.
\newblock \href {https://aclanthology.org/2024.bea-1.30} {Using large language
  models to assess young students{'} writing revisions}.
\newblock In \emph{Proceedings of the 19th Workshop on Innovative Use of NLP
  for Building Educational Applications (BEA 2024)}, pages 365--380, Mexico
  City, Mexico. Association for Computational Linguistics.

\bibitem[{Litman et~al.(2022)Litman, Afrin, Kashefi, Olshefski, Godley, and
  Hwa}]{litman2022argrewrite}
Diane Litman, Tazin Afrin, Omid Kashefi, Christopher Olshefski, Amanda Godley,
  and Rebecca Hwa. 2022.
\newblock \href {https://doi.org/10.1007/978-3-031-11644-5_52} {An automated
  writing evaluation system for supporting self-monitored revising}.
\newblock In \emph{Artificial Intelligence in Education: 23rd International
  Conference, AIED 2022, Durham, UK, July 27–31, 2022, Proceedings, Part I},
  page 581–587, Berlin, Heidelberg. Springer-Verlag.

\bibitem[{Liu and Zhu(2022)}]{liu2022bertalign}
Lei Liu and Min Zhu. 2022.
\newblock Bertalign: Improved word embedding-based sentence alignment for
  chinese--english parallel corpora of literary texts.
\newblock \emph{Digital Scholarship in the Humanities}.

\bibitem[{Liu et~al.(2023)Liu, Litman, Wang, Matsumura, and
  Correnti}]{liu-etal-2023-predicting}
Zhexiong Liu, Diane Litman, Elaine Wang, Lindsay Matsumura, and Richard
  Correnti. 2023.
\newblock \href {https://doi.org/10.18653/v1/2023.bea-1.24} {Predicting the
  quality of revisions in argumentative writing}.
\newblock In \emph{Proceedings of the 18th Workshop on Innovative Use of NLP
  for Building Educational Applications (BEA 2023)}, pages 275--287, Toronto,
  Canada. Association for Computational Linguistics.

\bibitem[{Mita et~al.(2024)Mita, Sakaguchi, Hagiwara, Mizumoto, Suzuki, and
  Inui}]{mita-etal-2024-towards}
Masato Mita, Keisuke Sakaguchi, Masato Hagiwara, Tomoya Mizumoto, Jun Suzuki,
  and Kentaro Inui. 2024.
\newblock \href {https://aclanthology.org/2024.bea-1.21} {Towards automated
  document revision: Grammatical error correction, fluency edits, and beyond}.
\newblock In \emph{Proceedings of the 19th Workshop on Innovative Use of NLP
  for Building Educational Applications (BEA 2024)}, pages 251--265, Mexico
  City, Mexico. Association for Computational Linguistics.

\bibitem[{Pennington et~al.(2014)Pennington, Socher, and
  Manning}]{pennington-etal-2014-glove}
Jeffrey Pennington, Richard Socher, and Christopher Manning. 2014.
\newblock \href {https://doi.org/10.3115/v1/D14-1162} {{G}lo{V}e: Global
  vectors for word representation}.
\newblock In \emph{Proceedings of the 2014 Conference on Empirical Methods in
  Natural Language Processing ({EMNLP})}, pages 1532--1543, Doha, Qatar.
  Association for Computational Linguistics.

\bibitem[{Rahimi et~al.(2017)Rahimi, Litman, Correnti, Wang, and
  Matsumura}]{rahimi2017assessing}
Zahra Rahimi, Diane Litman, Richard Correnti, Elaine Wang, and Lindsay~Clare
  Matsumura. 2017.
\newblock Assessing students’ use of evidence and organization in
  response-to-text writing: Using natural language processing for rubric-based
  automated scoring.
\newblock \emph{International Journal of Artificial Intelligence in Education},
  27(4):694--728.

\bibitem[{Roscoe et~al.(2013)Roscoe, Snow, and
  McNamara}]{10.1007/978-3-642-39112-5_27}
Rod~D. Roscoe, Erica~L. Snow, and Danielle~S. McNamara. 2013.
\newblock Feedback and revising in an intelligent tutoring system for writing
  strategies.
\newblock In \emph{Artificial Intelligence in Education}, pages 259--268,
  Berlin, Heidelberg. Springer Berlin Heidelberg.

\bibitem[{Sanh et~al.(2019)Sanh, Debut, Chaumond, and
  Wolf}]{sanh2019distilbert}
Victor Sanh, Lysandre Debut, Julien Chaumond, and Thomas Wolf. 2019.
\newblock \href
  {https://www.emc2-ai.org/assets/docs/neurips-19/emc2-neurips19-paper-33.pdf}
  {Distilbert, a distilled version of bert: smaller, faster, cheaper and
  lighter}.
\newblock In \emph{EMC2: Efficient Methods for Deep Learning Workshop at
  NeurIPS 2019}.

\bibitem[{Shibani et~al.(2018)Shibani, Knight, and
  Buckingham~Shum}]{shibani2018kb}
Antonette Shibani, Simon Knight, and Simon Buckingham~Shum. 2018.
\newblock Understanding revisions in student writing through revision graphs.
\newblock In \emph{International Conference on Artificial Intelligence in
  Education}, pages 332--336, Cham. Springer International Publishing.

\bibitem[{Spangher et~al.(2022)Spangher, Ren, May, and
  Peng}]{spangher-etal-2022-newsedits}
Alexander Spangher, Xiang Ren, Jonathan May, and Nanyun Peng. 2022.
\newblock \href {https://doi.org/10.18653/v1/2022.naacl-main.10}
  {{N}ews{E}dits: A news article revision dataset and a novel document-level
  reasoning challenge}.
\newblock In \emph{Proceedings of the 2022 Conference of the North American
  Chapter of the Association for Computational Linguistics: Human Language
  Technologies}, pages 127--157, Seattle, United States. Association for
  Computational Linguistics.

\bibitem[{Wang et~al.(2020)Wang, Matsumura, Correnti, Litman, Zhang, Howe,
  Magooda, and Quintana}]{wang2020eRevise}
Elaine~Lin Wang, Lindsay~Clare Matsumura, Richard Correnti, Diane Litman,
  Haoran Zhang, Emily Howe, Ahmed Magooda, and Rafael Quintana. 2020.
\newblock erevis(ing): Students’ revision of text evidence use in an
  automated writing evaluation system.
\newblock \emph{Assessing Writing}, 44:100449.

\bibitem[{Wilson et~al.(2021)Wilson, Huang, Palermo, Beard, and
  MacArthur}]{wilson2021automated}
Joshua Wilson, Yue Huang, Corey Palermo, Gaysha Beard, and Charles~A MacArthur.
  2021.
\newblock Automated feedback and automated scoring in the elementary grades:
  Usage, attitudes, and associations with writing outcomes in a districtwide
  implementation of mi write.
\newblock \emph{International Journal of Artificial Intelligence in Education},
  31(2):234--276.

\bibitem[{Zhang et~al.(2017)Zhang, Hashemi, Hwa, and
  Litman}]{zhang-etal-2017-corpus}
Fan Zhang, Homa~B. Hashemi, Rebecca Hwa, and Diane Litman. 2017.
\newblock \href {https://doi.org/10.18653/v1/P17-1144} {A corpus of annotated
  revisions for studying argumentative writing}.
\newblock In \emph{Proceedings of the 55th Annual Meeting of the Association
  for Computational Linguistics (Volume 1: Long Papers)}, pages 1568--1578,
  Vancouver, Canada. Association for Computational Linguistics.

\bibitem[{Zhang et~al.(2016)Zhang, Hwa, Litman, and
  Hashemi}]{zhang2016argrewrite}
Fan Zhang, Rebecca Hwa, Diane Litman, and Homa~B Hashemi. 2016.
\newblock Argrewrite: A web-based revision assistant for argumentative
  writings.
\newblock In \emph{Proceedings of the 2016 conference of the north american
  chapter of the association for computational linguistics: Demonstrations},
  pages 37--41.

\bibitem[{Zhang and Litman(2020)}]{zhang-litman-2020-automated}
Haoran Zhang and Diane Litman. 2020.
\newblock \href {https://doi.org/10.18653/v1/2020.acl-main.759} {Automated
  topical component extraction using neural network attention scores from
  source-based essay scoring}.
\newblock In \emph{Proceedings of the 58th Annual Meeting of the Association
  for Computational Linguistics}, pages 8569--8584, Online. Association for
  Computational Linguistics.

\bibitem[{Zhang and Litman(2021)}]{zhang-litman-2021-essay}
Haoran Zhang and Diane Litman. 2021.
\newblock \href {https://aclanthology.org/2021.bea-1.9} {Essay quality signals
  as weak supervision for source-based essay scoring}.
\newblock In \emph{Proceedings of the 16th Workshop on Innovative Use of NLP
  for Building Educational Applications}, pages 85--96, Online. Association for
  Computational Linguistics.

\bibitem[{Zhang et~al.(2019)Zhang, Magooda, Litman, Correnti, Wang, Matsmura,
  Howe, and Quintana}]{zhang2019erevise}
Haoran Zhang, Ahmed Magooda, Diane Litman, Richard Correnti, Elaine Wang,
  LC~Matsmura, Emily Howe, and Rafael Quintana. 2019.
\newblock erevise: Using natural language processing to provide formative
  feedback on text evidence usage in student writing.
\newblock In \emph{Proceedings of the AAAI Conference on Artificial
  Intelligence}, volume~33, pages 9619--9625.

\bibitem[{Ziegenbein et~al.(2024)Ziegenbein, Skitalinskaya, Bayat~Makou, and
  Wachsmuth}]{ziegenbein-etal-2024-llm}
Timon Ziegenbein, Gabriella Skitalinskaya, Alireza Bayat~Makou, and Henning
  Wachsmuth. 2024.
\newblock \href {https://doi.org/10.18653/v1/2024.acl-long.244} {{LLM}-based
  rewriting of inappropriate argumentation using reinforcement learning from
  machine feedback}.
\newblock In \emph{Proceedings of the 62nd Annual Meeting of the Association
  for Computational Linguistics (Volume 1: Long Papers)}, pages 4455--4476,
  Bangkok, Thailand. Association for Computational Linguistics.

\end{thebibliography}
\newpage
\appendix
\label{sec:appendix}
\section{MVP Article}
\label{sec:mvp_article}
{\it
\begin{center}
    {A Brighter Future}
\end{center}
\begin{center}
    Hannah Sachs
\end{center}

The unpaved dirt road made our car jump as we traveled to the Millennium Village in Sauri (sah-ooh-ree), Kenya. We passed the market where women sat on the dusty ground selling bananas. Little kids were wrapped in cloth on their mothers' backs, or running around in bare feet and tattered clothing. When we reached the village, we walked to the Bar Sauri Primary School to meet the people. Welcoming music and singing had almost everyone dancing. We joined the dancing and clapped along to the joyful, lively music.

The year was 2014, the first time I had ever been to Sauri. With the help of the Millennium Villages project, the place would change dramatically in the coming years. The Millennium Villages project was created to help reach the Millennium Development Goals. The plan is to get people out of poverty, assure them access to health care and help them stabilize the economy and quality of life in their communities. Villages get technical advice and practical items, such as fertilizer, medicine and school supplies. Local leaders take it from there. The goals are supposed to be met by 2025; some other targets are set for 2035. We are halfway to 2035, and the world is capable of meeting these goals. But our first glimpse of Sauri showed us that there was plenty of work to do.

\begin{center}
    {The Fight for Better Health}
\end{center}

On that day in 2014, we followed the village leaders into Yala Sub-District Hospital. It was not in good shape. There were three kids to a bed and two adults to a bed. The rooms were packed with patients who probably would not receive treatment, either because the hospital did not have it or the patients could not afford it. There was no doctor, only a clinical officer running the hospital. There was no running water or electricity. It is hard for me to see people sick with preventable diseases people who are near death when they shouldn't have to be. I just get scared and sad.

Malaria (mah-lair-eeh-ah) is one disease, common in Africa, that is preventable and treatable. Mosquitoes carry malaria, and infect people by biting them. Kids can die from it easily, and adults get very sick. Mosquitoes that carry malaria come at night. A bed net, treated with chemicals that last for five years, keeps malarial mosquitoes away from sleeping people. Each net costs \$5. There are some cheap medicines to get rid of malaria too. The solutions are simple, yet 20,000 kids die from the disease each day. So sad, and so illogical. Bed nets could save millions of lives.

\begin{center}
{Water, Fertilizer, Knowledge}
\end{center}

We walked over to see the farmers. Their crops were dying because they could not afford the necessary fertilizer and irrigation. Time and again, a family will plant seeds only to have an outcome of poor crops because of lack of fertilizer and water. Each year, the farmers worry: Will they harvest enough food to feed the whole family? Will their kids go hungry and become sick?

Many kids in Sauri did not attend school because their parents could not afford school fees. Some kids are needed to help with chores, such as fetching water and wood. In 2014, the schools had minimal supplies like books, paper and pencils, but the students wanted to learn. All of them worked hard with the few supplies they had. It was hard for them to concentrate, though, as there was no midday meal. By the end of the day, kids didn't have any energy.

\begin{center}
{A Better Life-2018}
\end{center}

The people of Sauri have made amazing progress in just four years. The Yala Sub-District Hospital has medicine, free of charge, for all of the most common diseases. Water is connected to the hospital, which also has a generator for electricity. Bed nets are used in every sleeping site in Sauri. The hunger crisis has been addressed with fertilizer and seeds, as well as the tools needed to maintain the food supply. There are no school fees, and the school now serves lunch for the students. The attendance rate is way up.

Dramatic changes have occurred in 80 villages across sub-Saharan Africa. The progress is encouraging to supporters of the Millennium Villages project. There are many solutions to the problems that keep people impoverished. What it will really take is for the world to work together to change poverty-stricken areas for good. When my kids are my age, I want this kind of poverty to be a thing of history. It will not be an easy task. But Sauri's progress shows us all that winning the fight against poverty is achievable in our lifetime.

\begin{center}
{Essay Prompt}
\end{center}

The author described how the quality of life can be improved by the Millennium Villages project in Sauri, Kenya. Based on the article, did the author convince you that “winning the fight against poverty is achievable in our lifetime”? Explain why or why not with 3 to 4 examples from the text to support your answer. \it}

\begin{table*}[t]
\begin{center}
\small
\begin{tabular}{|p{0.04\linewidth}|p{0.14\linewidth}|p{0.11\linewidth}|p{0.6\linewidth}|}
\hline \bf Level & \bf Assessment & \bf Name & \bf Feedback  \\ \hline
EF1& NPE less or equal to $\alpha$ & Needs more evidence & 
\textbullet Adding more evidence would make your argument even more convincing.

\textbullet Reread the highlighted portions of the article to choose more evidence (only use if texts were highlighted). 
\\ \hline
EF2& NPE greater than $\alpha$, SPC less or equal to $\beta$ & Needs more specific details & 
\textbullet Adding more details will help your reader better understand your ideas. This will make your argument even more convincing. 

\textbullet When you revise your essay, make sure to add more details for each piece of evidence you use.
\\ \hline
EF3& NPE greater than $\alpha$, SPC greater than $\beta$ & Needs more explanation & 
\textbullet Having evidence is important, but you need to help your reader understand how the evidence you chose supports your argument.

\textbullet When you revise your essay, focus on explaining how each piece of evidence you used connects to your idea. 

\textbullet Give a detailed and clear explanation of how the evidence supports your argument. 

\textbullet Tie the evidence not only to the point you are making within a paragraph, but to your overall argument. 

\\ \hline

\end{tabular}
\end{center}
\caption{\label{tab:feedbacks-evidence} Selecting evidence use feedback levels using NLP-based assessment.}
\vspace{-1em}
\end{table*}

\section{SPACE Article}
\label{sec:space_article}
{\it
\begin{center}
    {The Importance of Space Exploration}
\end{center}

Is space exploration really worthwhile when so much needs to be done on Earth? This is a question that has been asked for decades and requires serious thought.

\begin{center}
    {Arguments Against Space Exploration}
\end{center}

The arguments against space exploration stem from a belief that the money spent could be used differently – to improve people’s lives. In 1953, President Eisenhower captured this viewpoint. He opposed the space program, saying that each rocket fired was a theft from citizens who suffer from hunger and poverty. Indeed, over 46.2 million mericans (15\%) live in Poverty. Nearly half of all Americans also have trouble paying for basic needs such as housing, food, or medicine at some point in their lives.

In other countries people are dying because they do not have access to clean water, medical care or simple solutions that prevent the spread of diseases. For example, malaria, a disease spread by mosquito bites kills over 3,000 African children every day. This could be prevented simply by hanging large nets over children’s beds. This would protect people from being bitten as they sleep. These nets cost only \$5 each, but most families at risk for malaria cannot afford them.

It is not just people that need help. The Earth is suffering also. Many scientists believe that pollution from burning fossil fuels (gasoline and oil) is destroying our air and oceans. We need new, cleaner forms of energy to power cars, homes, and factories. A program to develop clean energy could be viewed as a worthy investment. Maybe exploring space should not be a priority when there is so much that needs to be done on Earth. In 2012, the United States spent 19 billion dollars for space exploration. Some people think that money should be spent to help heal the people and the Earth.

\begin{center}
    {Benefits of Space Exploration}
\end{center}

Those in favor of space exploration argue that 19 billion dollars is not too much. Nineteen billion was only 1.2\% of the total national budget. This is a tiny amount compared to the 670 billion spent that year on national defense (26.3\% of the national budget). It is even less than the 70 billion the federal government spent on education (48\% of the budget).

Another reason to spend money on space exploration is that it has led to benefits in many fields. One such field is medicine. Before NASA could send astronauts into space, scientists needed to find ways to monitor their health under stressful conditions. NASA wanted to make sure the astronauts could survive the harsh conditions of launch and reentry. The scientists developed medical instruments to monitor body functions. They also learned a lot about how the human body reacts to stress. In rising to meet the challenges of space exploration, NASA scientists have developed other innovations that improve our lives. These include better exercise machines and airplanes, and more accurate weather forecasting. All of these innovations resulted from technologies that engineers developed to make space travel possible.

Even the problems of hunger and poverty can be tackled by space exploration. Satellites that circle Earth can monitor land and the atmosphere. They can track and measure the conditions of crops, soil, and rainfall. We can use this information to improve the way we produce and distribute food. This will enable us to provide more food at a lower price to people who need it. When we explore space, we are also helping to solve serious problems on Earth.

Beyond providing us with inventions, space exploration is important for the challenge it provides and the motivation to bring out the best in ourselves. Often we make progress in solving difficult problems by first setting challenging goals, which inspire innovative work. Space exploration is important because it can motivate beneficial competition among nations. Imagine how much human suffering can be avoided if nations competed with planet-exploring spacceships instead of bomb-dropping airplanes. We saw an example of this during the 1960s. During the Cold War, the United States and Russia competed in a race to explore space. They each wanted to be the first to land a spacecraft on the moon and visit other planets. The National Academy of Science says that this competition led to significant investments and improvements in American education, especially in math and science. This shows that by looking outward into space, we also improve life here on Earth.

All this brings us back to the question: Should we explore space when there is so much that needs to be done on Earth? It is true that we have many serious problems to deal with on Earth. However, space exploration is not at odds with solving human problems. In fact, it may even help find solutions. Space exploration leads to long-term benefits that more than justify the immediate cost.

\begin{center}
    {Essay Prompt}
\end{center}
Consider the reasons given in the article for why we should and should not fund space exploration. Did the author convince you that “space exploration leads to long-term benefits” that justify the cost? Give reasons for your answer. Support your reasons with 3 to 4 pieces of evidence from the text.

\it}

\section{Implementation Detail} 
\label{appendix:implementation} 
The revision classifiers are implemented with PyTorch\footnote{\url{https://pytorch.org/}}, Huggingface\footnote{\url{https://huggingface.co/}}, and Azure OpenAI API\footnote{\url{https://azure.microsoft.com/en-us/}} for ChatGPT-3.5-turbo. We use pretrained DistilRoBERTa-base and BERT-base models from Huggingface as text encoders and multilayer perceptrons as classifier layers. The classifiers are optimized using cross-entropy loss with Adam optimizer on a GeForce RTX 3090 GPU. We set the batch size as 16 and the learning rate as 5e-5 with 5-fold cross-validation, where 80\% for training and 20\% for parameter tuning using the corpora from prior work. We use NLTK\footnote{\url{https://www.nltk.org/}} and sentence-splitter\footnote{\url{https://pypi.org/project/sentence-splitter/}} for text preprocessing. 

\begin{table*}[t]
\begin{center}
\small
\begin{tabular}{|p{0.04\linewidth}|p{0.15\linewidth}|p{0.11\linewidth}|p{0.6\linewidth}|}
\hline \bf Level & \bf Assessment & \bf Name & \bf Feedback                                                                                        \\ \hline
RF1& No content and no surface revisions or all deletions                  & No attempt                                     & \textbullet When writers revise, they generally add more content. This often makes their essays longer.

\textbullet This time when you revise your essay, focus on adding more evidence.                                                                                                                                                                                                                                               \\ \hline
RF2& No content revision                                                   & surface revision                               & \textbullet When you revised your essay, it looks like you edited your writing to be clearer and easier for a reader to understand. 

\textbullet Revising is different from editing. When writers revise their essays, they generally add more content. This often makes their essays longer. 

\textbullet This time when you revise your essay, focus on adding more evidence.                                           \\ \hline
RF3& No NPE change; Added sentences contain topic words already used       & Repeats evidence                               & \textbullet When you revised your essay, it looks like you added in evidence that was very similar to the evidence you had included before. 

\textbullet When writers revise, they generally add new content to their essays.                                                                                                                                                                                                           \\ \hline
RF4& No NPE change; Added sentences contain no topic words                 & Added evidence but not text based              & \textbullet When you revised your essay, it looks like you added more information about your thinking but did not include new information from the article. 

\textbullet When writers revise their text-based essays, they generally add new content from the text and delete content that is not based on the text.                                                                                                                    \\ \hline
RF5& NPE change (new NPE \textgreater old NPE) and SPC changes less or equal to $\gamma$ & Added evidence but vague or general            & \textbullet When you revised your essay, it looks like you followed the suggestion to add more evidence. Great job! 

\textbullet When writers revise, they don’t just add more information. They also add more details to the information they already have in their essay. This often makes their essays longer.                                                                                                                       \\ \hline
RF6& SPC changes less or equal to $\gamma$ and added sentences contain no SPC words      & Added evidence successfully                    & \textbullet When you revised your essay, it looks like you followed the suggestion to add more evidence. Great job! 

\textbullet Paying attention to feedback is how people become stronger writers.                                                                                                                                                                                                                                    \\ \hline
RF7& SPC changes greater than $\gamma$ (new SPC \textgreater old SPC)             & Added specific details successfully            & 
\textbullet When you revised your essay, it looks like you followed the suggestion to add more details to your essay. Great job! 

\textbullet Paying attention to feedback is how people become stronger writers.                                                                                                                                                                                                                       \\ \hline
RF8& No reasoning revision                                                 & Successful evidence but no reasoning attempt   & \textbullet When you revised your essay, it looks like you may have focused on something other than explaining your evidence. 

\textbullet Revising the explanation or reasoning part of an essay is hard to do! When writers revise for this, they make sure that after providing a piece of evidence, they say something that connects it to their argument. The explanation should not just restate the evidence in different words. \\ \hline
RF9& Unsuccessful reasoning revision                                       & Successful evidence but unsuccessful reasoning & \textbullet When you revised your essay, it looks like you may have focused on something other than explaining your evidence. 

\textbullet Revising the explanation or reasoning part of an essay is hard to do! When writers revise for this, they make sure that after providing a piece of evidence, they say something that connects it to their argument. The explanation should not just restate the evidence in different words. \\ \hline
RF10& Successful reasoning revision                                        & successful evidence and successful reasoning   & \textbullet When you revised your essay, it looks like you followed the suggestion to explain your evidence and how it connects to your claim. Great job! 

\textbullet Paying attention to feedback is how people become stronger writers.                                                                                                                                                                                              \\ \hline
\end{tabular}
\end{center}
\caption{\label{tab:feedbacks-revision} Selecting revision feedback levels using NLP-based assessments.}
\vspace{-1em}
\end{table*}

\section{Evidence Use Feedback Selection}
\label{sec:evidence_use_feedback_selection}
\begin{figure*}[t]
    \centering
    \includegraphics[width=0.9\linewidth]{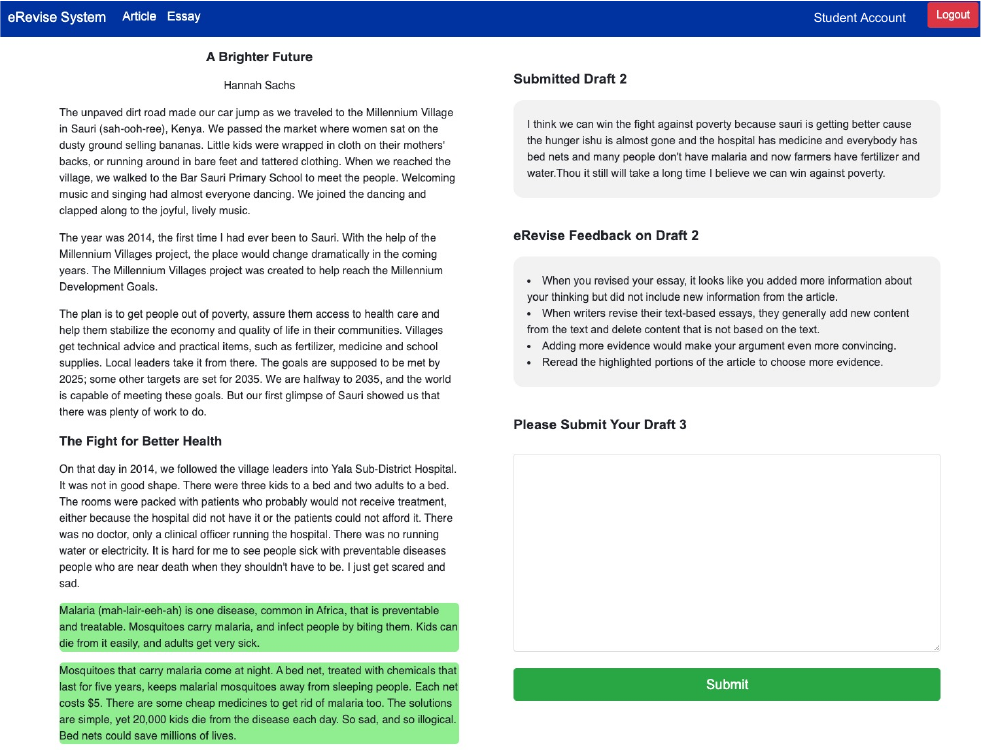}
    \caption{eRevise+RF System Student Interface: Main Page.}
    \label{fig:interface-student}
\end{figure*}

The AES system provides 3 levels of Evidence Use Feedback. An essay receives EF1 if its NPE score is less or equal to $\alpha$, which indicates no more than half of the topics are mentioned by the essay\footnote{We highlight text portions with a green background in the frontend (Figure~\ref{fig:interface-student}) to help students focus on missing topics if NPE covers no more than half of the topics.}; it receives EF2 when the NPE score is greater than $\alpha$ and SPC is less than or equal to $\beta$, which indicates the essay discusses several topics but details are still missing; it receives EF3 when the NPE score is greater than $\alpha$ and SPC is greater than $\beta$, which indicates that since the completeness and specificity of evidence are satisfactory, the student should focus on explaining how each piece of evidence is connected to a claim. The threshold $\alpha$ for NPE is set to half of the topics discussed in the relevant RTA source article, and $\beta$ for SPC is set to half of the example categories. Specifically, $\alpha$ and $\beta$ are set to 2 and 4. The feedback messages and how they are selected are shown in Table~\ref{tab:feedbacks-evidence}. 

\section{Revision Feedback Selection}
\label{sec:revision_feedback_selection}
RF addresses 10 levels of revisions organized on a revision feedback tree in Figure~\ref{fig:tree}. The system uses scoring indicators (NPE and SPC) and revision classifiers (RC-Content, RC-Evidence, and RC-Success) to select RF from the paths on the feedback tree. Specifically, \textit{Path RF-RF1}: In the case a student makes no revision when all aligned revision pairs are identical, the feedback is RF1. \textit{Path RF-RF2}: If a student makes only surface revisions as predicted by RC-Content, RF2 is used. In the cases where the last draft had EF1, the RF has 4 sub-paths. \textit{Path RF-EF1-RF3}: RF3 is given if there are no NPE changes between the drafts, and the newly added sentences contain topic words that already exist. \textit{Path RF-EF1-RF4}: RF4 is used if NPE has no increases between drafts, and added sentences contain no topic words. \textit{Path RF-EF1-RF5}: It provides RF5 if NPE increases, but the increment of SPC is less than an expert-designed threshold $\gamma$ ($\gamma$ is set to 2). \textit{Path RF-EF1-RF6}: RF6 recognizes the student's successful revisions in response to EF1 (completeness), thus needs to focus on EF2 (specificity). In the cases where EF on the last draft had EF2, the RF has 3 sub-paths. \textit{Path RF-EF2-RF4}: RF4 is used if SPC increment is less than $\gamma$, and added sentences contain no SPC words\footnote{We use expert-crafted keywords for specific RTA article examples as in~\citet{rahimi2017assessing}.}. \textit{Path RF-EF2-RF5}: RF5 is used if SPC increment is less than $\gamma$ but contains SPC words. \textit{Path RF-EF2-RF7}: RF7 is selected when the revision successfully addressed EF2. In the cases where EF on the last draft was EF3, RF has 3 sub-paths. \textit{Path RF-EF3-RF8}: RF8 is used if there is no reasoning revision pair predicted by RC-Evidence. \textit{Path RF-EF3-RF9}: RF9 is used when an unsuccessful revision label is predicted by RC-Evidence and RC-Success. \textit{Path RF-EF3-RF10}: RF10 is provided if not RF8 or RF9. The detailed messages and how they are selected are shown in Table~\ref{tab:feedbacks-revision}.

\begin{table*}[!htb]
\tiny
\centering
\begin{adjustbox}{width=2\columnwidth}
\begin{tabular}{p{0.01\textwidth}p{0.15\textwidth}p{0.15\textwidth}p{0.06\textwidth}p{0.06\textwidth}p{0.10\textwidth}p{0.09\textwidth}}		
\hline
\textbf{ID} & \textbf{Original Draft Sentence} & \textbf{Revised Draft Sentence} & \textbf{Revision Action} & \textbf{Revision Type} & \textbf{Revision Purpose} & \textbf{Quality Label} \\ \midrule
0           & In the article "a brighter future", the author proved to me that it is possible to end poverty in a lifetime.        & In the article "a brighter future", the author proved to me that it is possible to end poverty in a lifetime.                                           & N/A               & N/A                    & N/A                       & N/A                    \\ \hline
1           & She did this by showing how other countries helped them, how they got more food, and how their water supply went up. & She did this by showing how when they worked together they made medication more affordable (if not free), got more food, and got their water supply up. & Modify            & Content                & Relevant Evidence         & Successful             \\ \hline
2           & ...                                                                                                                  & ...                                                                                                                                                     & ...               & ...                    & ...                       & ...                    \\ \hline
3           & Although some countries are having wars right now, we are still one big team.                                        &                                                                                                                                                         & Delete            & Content                & not Text-based Evidence   & Successful             \\ \hline
4           &                                                                                                                      & First, they did it by making medication more affordable.                                                                                                & Add               & Content                & Relevant Evidence         & Successful             \\ \hline
5           & ...                                                                                                                  & ...                                                                                                                                                     & ...               & ...                    & ...                       & ...                    \\ \hline
6           &                                                                                                                      & It is most common in Africa.                                                                                                                            & Add               & Content                & not Text-based Evidence   & Unsuccessful           \\ \hline
7           & I know this because the author said, "water is connected to the hospital.                                            & I know this because the author said, "water is connected to the hospital."                                                                              & Modify            & Surface                & N/A                       & N/A                    \\ \hline
8           & Water is critical to life, so if there is a shortage then no wonder people were dying.                               & Water is critical to life, so if there is a shortage then no wonder people were dying.                                                                  & N/A               & N/A                    & N/A                       & N/A                    \\ \hline
9           & ...                                                                                                                  & All of these reasons make it clear that if you put your mind to something, you can achieve it.                                                          & Add               & Content                & LCE Reasoning             & Successful             \\ \hline
10          & ...                                                                                                                  & ...                                                                                                                                                     & ...               & ...                    & ...                       & ...                    \\ \hline
11          & Life is hard, but dying is harder.                                                                                   & Life is hard, but you don't have to work alone on hard things.                                                                                          & Modify            & Content                & Commentary Reasoning      & Unsuccessful         \\ 
\hline
\end{tabular}
\end{adjustbox}
\caption{Example of revision annotation between two essay drafts. The sentence alignment (revision action) and classification (revision type) were first done by the system and then manually justified and corrected by annotators. The revision purpose and quality label are annotated based on the RER scheme~\cite{afrin2020RER,liu-etal-2023-predicting}.}
\label{table:data-example-alignment}
\end{table*}

\begin{table*}[th]
\centering
\small
\begin{tabular}{c|ccc|ccc|ccc}
\hline
        & \multicolumn{3}{c}{RC-Content} & \multicolumn{3}{c}{RC-Evidence} & \multicolumn{3}{c}{RC-Success} \\ \hline
        & Precision   & Recall   & F1-Score    & Precision   & Recall   & F1-Score     & Precision   & Recall   & F1-Score    \\ \hline
Grade 4 & 0.90        & 0.92     & 0.91  & 0.85        & 0.53     & 0.65   & 0.64        & 0.87     & 0.74  \\
Grade 5 & 0.99        & 0.97     & 0.98  & 0.78        & 0.59     & 0.67   & 0.78        & 0.84     & 0.81  \\
Grade 6 & 0.93        & 0.97     & 0.95  & 0.89        & 0.55     & 0.68   & 0.47        & 1.00     & 0.64  \\
Grade 7 & 0.97        & 0.99     & 0.98  & 0.77        & 0.59     & 0.67   & 0.43        & 0.93     & 0.59  \\
Grade 8 & 0.98        & 0.98     & 0.98  & 0.56        & 0.56     & 0.56   & 0.65        & 0.90     & 0.76  \\ \hline
Overall & 0.96        & 0.97     & 0.96  & 0.78        & 0.57     & 0.66   & 0.58        & 0.88     & 0.70 \\ \hline
\end{tabular}
\caption{The grade-level evaluation of revision classifiers based on predicted values vs. human annotations. The RC-Content classifier has good performance on all grades, while RC-Evidence has relatively lower results in grade 8 and the precision in grades 6 and 7 are relatively low in RC-Success.}
\label{tab:f1-recall-precision-classifier}
\end{table*}

\begin{table*}[t]
\small
\centering
\begin{tabular}{|ccc|ccc|ccc|}
\hline
\multicolumn{3}{|c|}{RC-Content}                                       & \multicolumn{3}{c|}{RC-Evidence}                                           & \multicolumn{3}{c|}{RC-Success}                                           \\ \hline
\multicolumn{1}{|c|}{}        & \multicolumn{1}{c|}{Surface} & Content & \multicolumn{1}{c|}{}          & \multicolumn{1}{c|}{Reasoning} & Evidence & \multicolumn{1}{c|}{}          & \multicolumn{1}{c|}{Unsuccess} & Success \\ \hline
\multicolumn{1}{|c|}{Surface} & \multicolumn{1}{c|}{268}     & 50      & \multicolumn{1}{c|}{Reasoning} & \multicolumn{1}{c|}{318}       & 99       & \multicolumn{1}{c|}{Unsuccess} & \multicolumn{1}{c|}{100}       & 358     \\ \hline
\multicolumn{1}{|c|}{Content} & \multicolumn{1}{c|}{37}      & 1170    & \multicolumn{1}{c|}{Evidence}  & \multicolumn{1}{c|}{269}       & 360      & \multicolumn{1}{c|}{Success}   & \multicolumn{1}{c|}{67}        & 499     \\ \hline
\end{tabular}
\caption{The confusion matrices for revision classifiers based on human annotations (rows) vs. predicted values (columns). The RC-Evidence is likely to predict evidence as reasoning and RC-Success makes errors mostly on true unsuccessful revisions.}
\label{tab:confusion-classifier}
\end{table*}

\begin{table*}[th]
\small
\centering
\begin{tabular}{c|cccc|cccc}
\hline
\multirow{2}{*}{\begin{tabular}[c]{@{}c@{}}Evidence\\ Feedback\end{tabular}} & \multicolumn{4}{c|}{Draft1-Draft2}                                     & \multicolumn{4}{c}{Draft2-Draft3}                                      \\ \cline{2-9} 
                                                                             & NPE       & \multicolumn{1}{c|}{$\Delta$NPE} & SPC       & $\Delta$SPC & NPE       & \multicolumn{1}{c|}{$\Delta$NPE} & SPC       & $\Delta$SPC \\ \hline
EF1 (N=55)                                                                   & 1.05/2.71 & \multicolumn{1}{c|}{+158\%}      & 1.6/4.91  & +207\%      & 2.71/3.27 & \multicolumn{1}{c|}{+20.7\%}     & 4.91/6.60 & +34.4\%     \\
EF2 (N=49)                                                                   & 3.35/3.84 & \multicolumn{1}{c|}{+14.6\%}     & 5.20/6.67 & +28.3\%     & 3.84/4.02 & \multicolumn{1}{c|}{+4.69\%}     & 6.67/7.55 & +13.2\%     \\
EF3 (N=46)                                                                   & 5.02/5.20 & \multicolumn{1}{c|}{+3.59\%}     & 12.3/12.7 & +3.16\%     & 5.20/5.15 & \multicolumn{1}{c|}{-0.96\%}     & 12.7/12.9 & +1.34\%     \\ \hline
\end{tabular}
\caption{Predicted NPE and SPC changes between two SPACE essay drafts with respect to received evidence feedback levels on the old drafts. The number before the slash is the NPE and SPC on the old drafts and the latter is on the new drafts. $\Delta$ means the ratio of the changes from the old drafts to the new drafts. Both evidence feedback (from draft1 to draft2) and revision feedback (from draft2 to draft3) help students
improve their essays.}
\label{tab:change-npe-spc-SPACE}
\end{table*}

\begin{table*}[t]
    \centering
    \includegraphics[width=1\linewidth]{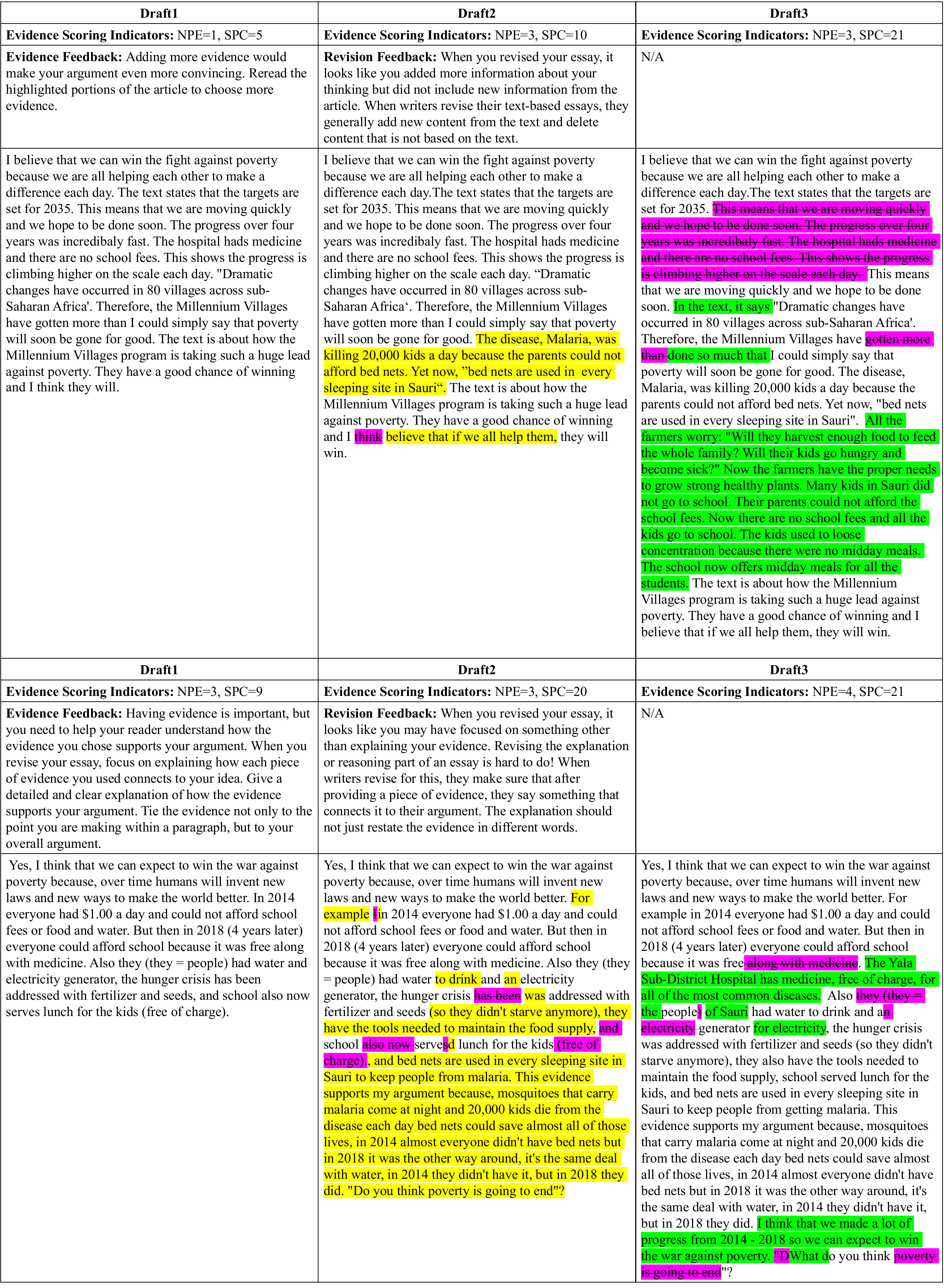}
    \caption{The examples of three essay drafts, revising from draft1 to draft2, then from draft2 to draft3. The purple marks deletion, yellow and green mark addition. The NPE and SPC scores have improved after revising from draft1 to draft2 based on the evidence feedback, and revising from draft2 to draft3 based on the revision feedback.}
    \label{fig:example-2}
    \vspace{-0.8em}
\end{table*}

\begin{table*}[t]
    \centering
    \includegraphics[width=1\linewidth]{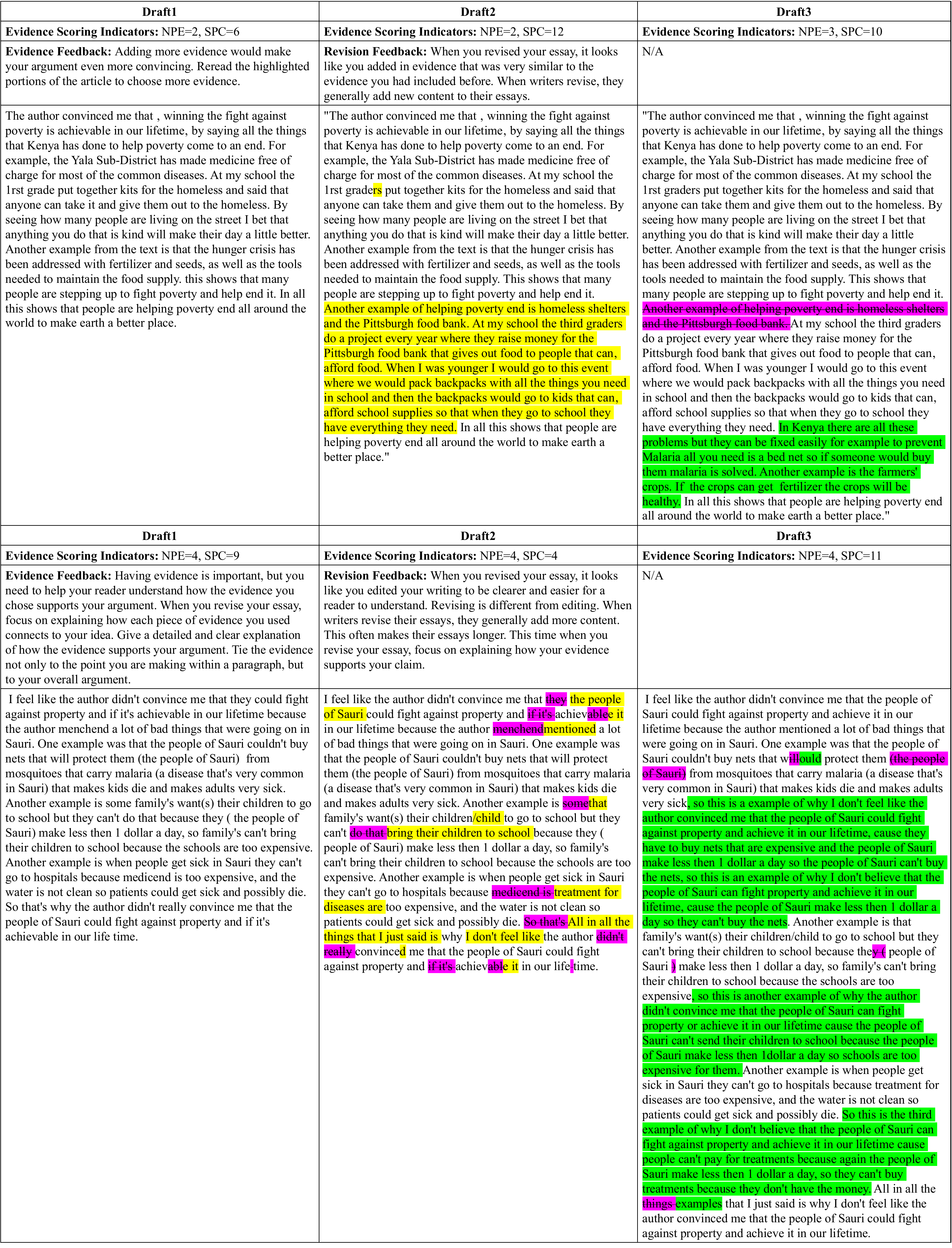}
    \caption{Additional examples of three essay drafts, revising from draft1 to draft2, then from draft2 to draft3. The purple marks deletion, yellow and green mark addition. The NPE and SPC scores have improved after revising from draft1 to draft2 based on the evidence feedback, and revising from draft2 to draft3 based on the revision feedback.}
    \label{fig:example-3}
    \vspace{-0.8em}
\end{table*}

\begin{figure*}[t]
    \centering
    \includegraphics[width=0.9\linewidth]{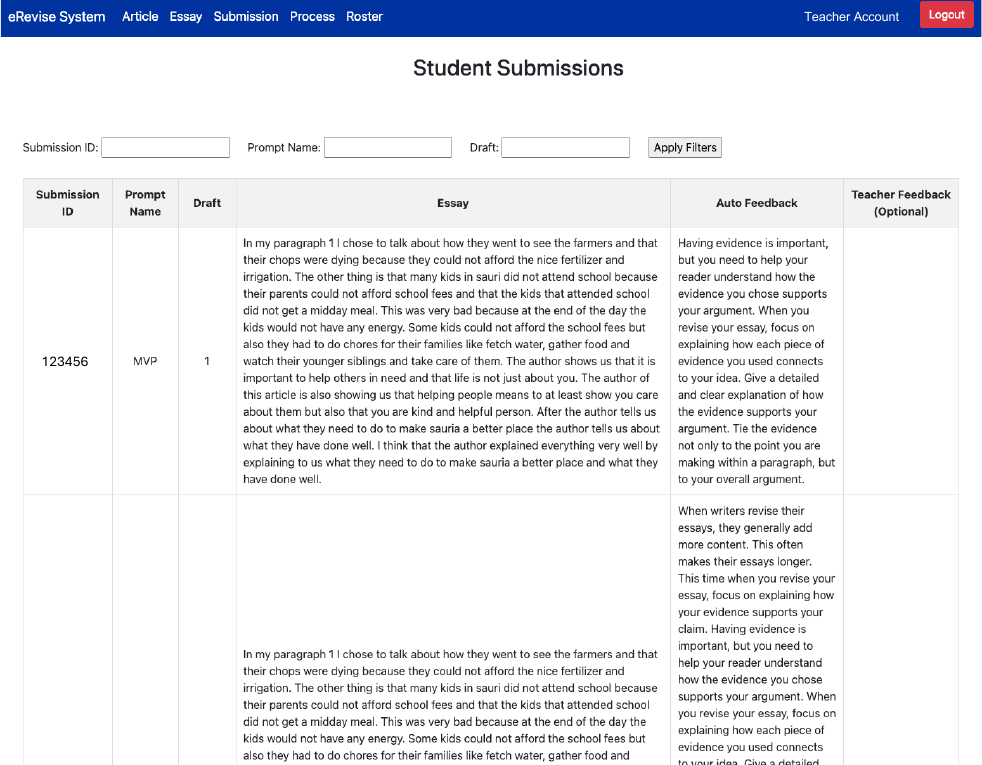}
    \caption{eRevise+RF System Teacher Interface: Submission Page.}
    \label{fig:interface-teacher}
\end{figure*}

\end{document}